\documentclass{article}

\usepackage{microtype}
\usepackage{graphicx}
\usepackage{booktabs} % for professional tables

\usepackage{hyperref}

\usepackage[accepted]{icml2025}

\usepackage{amsmath}
\usepackage{amssymb}
\usepackage{mathtools}
\usepackage{amsthm}

\usepackage[capitalize,noabbrev]{cleveref}

\theoremstyle{plain}

\theoremstyle{definition}

\theoremstyle{remark}

\usepackage[textsize=tiny]{todonotes}

\icmltitlerunning{\myname{}: Targeted Human Feedback for LLM Alignment}

\usepackage{enumitem}
\usepackage{multirow}
\usepackage{tcolorbox}
\usepackage{moreverb}
\usepackage{cprotect}
\usepackage{fancyvrb}
\usepackage{framed}
\usepackage{subcaption}

\newcommand{\myname}[0]{RLTHF}

\newcommand{\bbb}[1]{\noindent\textbf{#1}}

\begin{document}

\twocolumn[
\icmltitle{\myname{}: Targeted Human Feedback for LLM Alignment}



\icmlsetsymbol{intern}{*}

\begin{icmlauthorlist}
% \icmlauthor{Anonymous Authors}{}
\icmlauthor{Yifei Xu}{microsoft,ucla,intern}
\icmlauthor{Tusher Chakraborty}{microsoft}
\icmlauthor{Emre K\i c\i man}{microsoft}
\icmlauthor{Bibek Aryal}{microsoft}
\icmlauthor{Eduardo Rodrigues}{microsoft}
\icmlauthor{Srinagesh Sharma}{microsoft}
\icmlauthor{Roberto Estevao}{microsoft}
\icmlauthor{Maria Angels de Luis Balaguer}{microsoft}
\icmlauthor{Jessica Wolk}{microsoft}
\icmlauthor{Rafael Padilha}{microsoft}
\icmlauthor{Leonardo Nunes}{microsoft}
\icmlauthor{Shobana Balakrishnan}{microsoft}
\icmlauthor{Songwu Lu}{ucla}
\icmlauthor{Ranveer Chandra}{microsoft}
\end{icmlauthorlist}

\icmlaffiliation{microsoft}{Microsoft}
\icmlaffiliation{ucla}{University of California, Los Angeles}

\icmlcorrespondingauthor{Yifei Xu}{yxu@cs.ucla.edu}
\icmlcorrespondingauthor{Tusher Chakraborty}{tusher.chakraborty@microsoft.com}

\icmlkeywords{LLM, RLHF, Alignment, Machine Learning, ICML}

\vskip 0.3in
]



% \printfootnote{\textsuperscript{*}Work is done during an internship at Microsoft Research.}  % leave blank if no need to mention equal contribution
\printAffiliationsAndNotice{\textsuperscript{*}Work done during an internship at Microsoft.} % otherwise use the standard text.

\begin{abstract}

Fine-tuning large language models (LLMs) to align with user preferences is challenging due to the high cost of quality human annotations in Reinforcement Learning from Human Feedback (RLHF) and the generalizability limitations of AI Feedback. To address these challenges, we propose \myname{}, a human-AI hybrid framework that combines LLM-based initial alignment with selective human annotations to achieve full-human annotation alignment with minimal effort. \myname{} identifies hard-to-annotate samples mislabeled by LLMs using a reward model's reward distribution and iteratively enhances alignment by integrating strategic human corrections while leveraging LLM's correctly labeled samples. Evaluations on HH-RLHF and TL;DR datasets show that \myname{} reaches full-human annotation-level alignment with only 6-7\% of the human annotation effort. Furthermore, models trained on \myname{}'s curated datasets for downstream tasks outperform those trained on fully human-annotated datasets, underscoring the effectiveness of \myname{}.
\end{abstract}

\section{Introduction}
\label{sec:intro}
\begin{figure*}[t]
  \centering
  \includegraphics[width=0.85\textwidth]{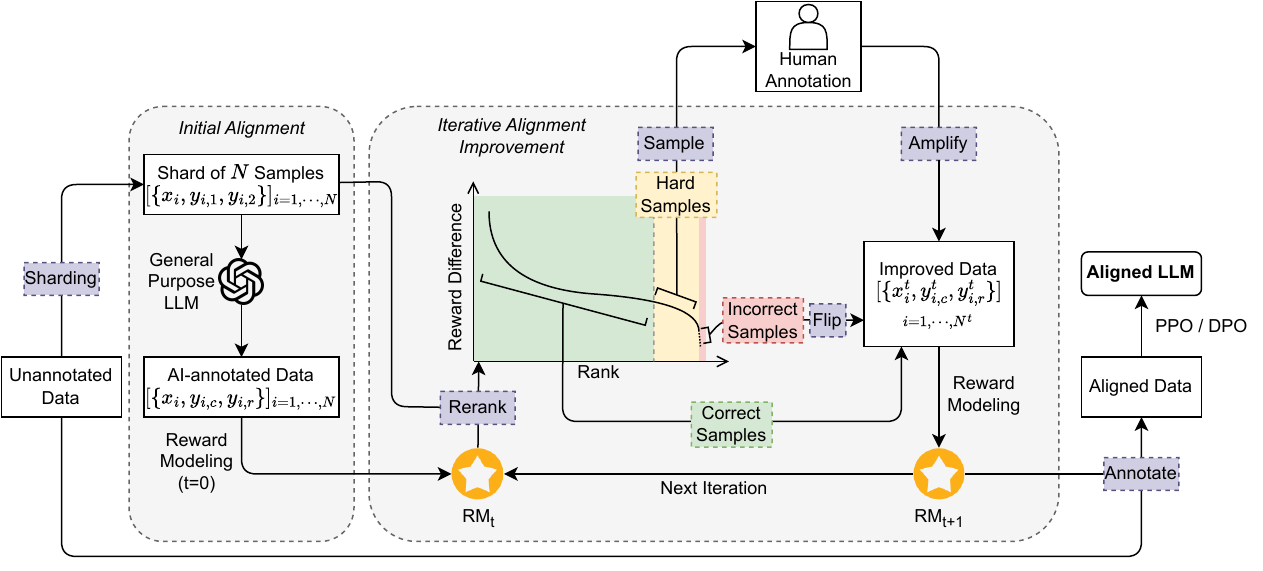}
  \caption{Overview of \myname{} process. \myname{} starts with coarse LLM alignment on the task. It then iteratively takes targeted human feedback and blending the same with sanitized LLM-annotations to reach the complete human alignment, leveraging reward distribution of a reward model in its training dataset.}
  \label{fig:overview}
  \vspace{-0.2in}
\end{figure*}

In recent years, large language models (LLMs) have demonstrated remarkable advancements, unlocking new possibilities across a wide range of applications~\cite{touvron2023llama, jiang2024mixtral, achiam2023gpt, team2023gemini}. As these models become more powerful, the focus has shifted toward customization, i.e., fine-tuning base models to better serve specific tasks and user needs~\cite{wei2021finetuned, li2023self}. Companies are increasingly investing in solutions built upon fine-tuned models, recognizing the value of adapting LLMs to align with end-user preferences, including intent, style, grounding, and compliance requirements~\cite{ft_service, m365, nuances, folio}. A key approach to achieving this alignment is Reinforcement Learning from Human Feedback (RLHF), which has emerged as a widely adopted technique in the literature for refining model behavior based on human feedback~\cite{bai2022training, stiennon2020learning, rafailov2024direct, wang2024secrets, ouyang2022training}.
% Large Language Models (LLMs) have achieved impressive success in a wide range of natural language processing (NLP) tasks. Recent advances in instruction tuning and alignment (e.g., RLHF/RLAIF~\cite{}) have further enabled these models to generate coherent and contextually appropriate responses to user queries. However, in enterprise-level applications, practitioners often need to fine-tune LLMs for \textit{custom tasks} such as domain-specific summarization, legal clause extraction, and domain question answering. 

The effectiveness of RLHF techniques heavily depends on high-quality human annotations, which are both costly and time-consuming to obtain~\cite{pang2023language, lee2023rlaif, wang2024secrets}. To mitigate this challenge, Reinforcement Learning from AI Feedback (RLAIF) has been introduced, leveraging LLMs to replace human annotators in the feedback loop~\cite{lee2023rlaif, leerlaif, bai2022constitutional}. While RLAIF can approximate human judgment to some extent, it is sensitive to factors such as prompt optimization, task complexity, model bias, generator-discriminator gap, and the capability of the judge model, limiting its ability to fully replicate human annotations~\cite{huang2024self, sharma2024critical, lee2023rlaif, zeng2024scaling, huang2023large}. Our evaluation also provides evidence of these limitations. Furthermore, the samples that challenge a judge model are often the ones most critical for adapting base models to specialized fine-tuning tasks~\cite{ethayarajh2024kto, yuan2024self, huang2023large}. The cost of human annotation is further exacerbated by privacy and security constraints that restrict fine-tuning service providers' access to an entire customer data corpus. In such cases, only subject matter experts (SMEs) within the customer organization have full visibility into the data, making it particularly difficult to optimize prompts effectively across the entire corpus, especially for hard-to-annotate samples.

To address these challenges, we propose Reinforcement Learning from Targeted Human Feedback (\myname{}), a human-AI hybrid solution that combines coarse initial alignment using general-purpose LLMs with the progressive integration of strategically selected human annotations to achieve annotation quality comparable to fully human-supervised approaches. \myname{} begins with an initial alignment stage, where a general-purpose LLM labels unlabeled data based on high-level instructions. While this approach effectively captures broader human alignment for easier data points, it often struggles with fine-grained nuances, leading to incorrect labeling. \myname{} automatically identifies these hard-to-annotate data points and directs human effort exclusively toward them. This targeted approach enables \myname{} to achieve the quality of fully human-annotated data while reducing the majority of human annotation effort.

To enable this efficient human-in-the-loop approach for achieving comprehensive human alignment, \myname{} introduces the following key technical contributions:

First, we introduce a concept that leverages the reward distribution of a reward model over its training dataset to capture the relative arrangement of samples based on rewarded features. This distributional property enables the identification of both potential annotation inaccuracies and the model’s confidence across the dataset. Specifically, we train a reward model on the LLM-labeled dataset to uncover clusters of hard-to-annotate samples that are \textit{highly} likely to be mislabeled by the LLM, as well as easy samples that are \textit{highly} likely to be correctly labeled.

Building on this concept, we propose an innovative iterative reward model training technique to achieve oracle-level human alignment in the dataset. In each iteration, \myname{} identifies highly probable mislabeled data points and rectifies the same using human annotations. Simultaneously, it detects clusters of samples that are very likely to be correctly labeled by the LLM and incorporates them with human-annotated data to construct a high-quality training set for the next iteration of reward model training. Throughout this process, \myname{} preserves data richness and maximizes the efficiency of human annotation investment through carefully controlled hyperparameters.

Finally, we evaluate \myname{} on two distinct preference datasets: HH-RLHF and TL;DR. Our results demonstrate that \myname{} achieves accuracy comparable to a fully human-annotated dataset while requiring only 6–7\% of the total human annotations. Furthermore, we conduct a comparative study by training models on downstream tasks using DPO~\cite{rafailov2024direct}. Remarkably, models trained with \myname{} even outperform those trained on fully human-annotated datasets, highlighting the impact of \myname{}'s meticulous data curation in enhancing model performance.

\section{Background and Related Work}
LLMs have demonstrated impressive performance across a wide spectrum of tasks~\cite{achiam2023gpt, dubey2024llama, team2023gemini}. Despite the progress, their performance on customized downstream tasks can be significantly optimized by supervised fine-tuning (SFT) with instruction and human-written responses pairs~\cite{chung2024scaling, thoppilan2022lamda}. Reinforcement learning with preference data has further shown success due to the easier-to-collect data form~\cite{ouyang2022training, stiennon2020learning, lee2023rlaif}. Representative methods include Proximal Policy Optimization
(PPO)~\cite{schulman2017proximal}, which optimizes the LLM with a separate reward model, and Direct Preference Optimization (DPO)~\cite{rafailov2024direct}, which directly learns from the preference data. Although an easier data collection is available, these methods still largely rely on the richness and quality of the preference data~\cite{xu2024dpo,zheng2023secrets, wang2024secrets}.

\vspace{-0.1in}

% Provide a very short background on fine-tuning, specifically fine-tuning oriented services. Talk about why quality data is important there, especially human preference is important. Then, describe the corresponding techniques and their limiotations. 
\subsection{Alignment with External Feedback}
% \begin{itemize}
%     \item RLHF: gold standard, but many human effort wasted on what LLM already knows
%     \item RLAIF: bounded by the capability of the feedback LLM, stuggling with unseen data/tasks
%     \item \myname  strategically direct human effort to unseen/hard high-value tasks, effectively use minimum human effort to achieve maximun alignment
\vspace{-0.05in}
Human feedback is regarded as the golden standard in LLM alignment. However, reinforcement learning from human feedback (RLHF)~\cite{ouyang2022training, stiennon2020learning, kopf2024openassistant} typically incorporates heavy and expensive crowdsourcing efforts or expert annotations to guarantee data diversity and richness. To relieve the reliance on human effort, reinforcement learning with AI feedback (RLAIF)~\cite{lee2023rlaif, bai2022constitutional} provides an alternative that collects feedback from stronger LLMs instead of humans. On the other hand, this method is limited by the capability of the stronger LLM annotators~\cite{huang2024self, sharma2024critical, lee2023rlaif} especially for customized tasks, and suffers from their intrinsic biases~\cite{zheng2023judging}.
In this paper, we take advantage of RLAIF to establish an initial alignment and strategically incorporate human feedback to efficiently bring LLMs to the true alignment.

% \end{itemize}
\vspace{-0.1in}
\subsection{LLM Self-Improvement}
\vspace{-0.05in}
% \begin{itemize}
%     \item Self-Rewarding LM/SER: bounded by the intrinsic capability of the LLM itself, stuggling with unseen data/tasks
%     \item \myname introduces new human intelligence and break the upper bound of LLM in domain understanding
% \end{itemize}

To break the upper bound of LLMs, recent efforts have been devoted to enabling LLMs to self-improve. Self-Rewarding LMs~\cite{yuan2024self} and Math-shepherd~\cite{wang2024math} demonstrate the possibility of LLM self-improvement with reward signals from itself. SELF-ALIGN~\cite{sun2024principle} uses a carefully written set of principles to guide LLMs through self-improvement. SER~\cite{huang2024self} starts with only a fraction of human annotations to achieve full-annotation performance by progressively generating additional training data for itself. 
However, these methods still suffer from the intrinsic upper bound of LLMs and self-improvement is not guaranteed for customized tasks.
\myname{}, on the other hand, efficiently introduces human intelligence into the improvement process, thereby ensuring that the improvement is not bounded by LLMs' initial lack of domain understanding.

\section{Improving Human Alignment with \myname{}}
\label{sec:design:overview}

\myname{} enhances alignment with human in preference datasets used for training preference optimization techniques like DPO and PPO. It facilitates LLM training for various downstream tasks, including summarization, compliance, and grounding. Starting with an unlabeled preference dataset, \myname{} strategically integrates AI-generated labels with selective human feedback to maximize alignment while minimizing annotation effort. As illustrated in Figure~\ref{fig:overview}, \myname{} operates in three stages: 1) \textit{Initial alignment}, where an off-the-shelf LLM provides dataset labeling to establish a coarse task understanding, 2) \textit{Iterative alignment improvement}, which leverages reward distribution by a reward model (RM) to locate hard-to-annotate samples mislabeled by the LLM and rectify with selective human feedback while investing the correct LLM labels, 3) \textit{Transferring knowledge for downstream task}, where the curated preference dataset is fed into the DPO pipeline or the trained \myname{} reward model is integrated into the PPO pipeline. Find the corresponding pseudocode in Appendix~\ref{appendix:pseudocode}.

% Specially, for text extraction, \myname{} separates the task into two steps: Filtering step applies the initial alignment and progressive hard-problem resolution; and the learned filtering RM is re-purposed to provide high-quality extractions.

\subsection{Initial Alignment}
\label{sec:design:init}

This stage aims to establish an initial coarse alignment in the unlabeled dataset using a general-purpose LLM, which provides preference annotations for each unannotated sample. Prior research suggests that model selection here depends on task complexity relative to the model's capability~\cite{snell2024scaling}. While \myname{} is not found to be sensitive to the choice of model at this stage, a well-suited model can accelerate alignment convergence. The only assumption is that the general-purpose LLM possesses a basic understanding of the downstream task, enabling it to provide a rough initial alignment that serves as a seed for \myname{}.     

Our prompt for obtaining preference judgments from the LLM consists of three components: 1) task description, 2) preference judgment principles provided by the user, and 3) few-shot examples with optional chain-of-thought reasoning. The prompt templates are detailed in Appendix~\ref{appendix:prompt}. We do not perform explicit fine-grained prompt tuning, as full visibility into the data may be restricted when offering fine-tuning services to third-party customers. However, to ensure that the selected LLM with our prompt attains a rough level of alignment, we perform an eyes-off validation using strategic human feedback, as detailed in Section~\ref{sec:leveraging_reward_score}.   

As mentioned earlier, this AI-generated feedback is prone to errors due to factors such as model biases from pre-training data, task complexity, and prompt optimization, which is also evident in our evaluation. When our ultimate goal is to customize an existing model through fine-tuning to align with end-user preferences, we inherently assume that an off-the-shelf LLM lacks comprehensive alignment with the end-user. However, \myname{} builds upon the initial AI-provided alignment and systematically refines it in subsequent stages to achieve oracle-level human alignment.

\subsection{Iterative Alignment Improvement}
\label{sec:design:pref_learn}
In this stage, we refine the LLM-labeled preference dataset by iteratively training an RM with selective human annotations to enhance alignment. Before diving into the details of this process, we first establish the premise for RM.

\subsubsection{Reward Model}
 Given a labeled preference dataset $\mathcal{D}_{\mathbf{\Lambda}} = \{x_i, y_{i,c}, y_{i,r} \}$, where $i\in [N]$, $x_i$ is the prompt, $y_{i,c}$ and $y_{i,r}$ denote the chosen and rejected completions, respectively, as labeled according to the annotator's preference, $\mathbf{\Lambda}$. Here, if we represent the relative preference orientation of $i^{th}$ completion pair with $\lambda = [-1, +1 ]$, $\mathbf{\Lambda}$ is a $N$-dimensional vector consists of $[\lambda_i]_{i=1}^{N}$, meaning that flipping the preferences of all completion pairs results in $\mathcal{D}_{\mathbf{-\Lambda}}$. To train an RM on this dataset, we can formulate the probability distribution of $y_{i,c}$ being preferred over $y_{i,r}$ given $x_{i}$ as an input, following the Bradley-Terry (BT) model~\cite{david1963method}.
\begin{equation}
    P(x\succ y) = \sigma(r(x_i, y_{i,c}) - r(x_i, y_{i,r}))
\label{eq:rm_prob}
\end{equation}
where $\sigma(\cdot)$ denotes the sigmoid function and $r(\cdot)$ denotes the reward function. Assuming the existence of a true deterministic reward function, the goal is to train the RM to learn this function and predict the reward, $\hat{r}(x,y)$. The RM training can be framed as a binary classification problem~\cite{sun2024rethinking}, where a labeled pair of $\rho_{i,c}\coloneqq(x_i, y_{i,c})$ and  $\rho_{i,r}\coloneqq(x_i, y_{i,r})$ is passed to the model to predict the conditional class probability according to Eq.~\ref{eq:rm_prob}. This leads to the negative log-likelihood loss function for training.
\begin{equation}
    \mathcal{L(\hat{\text r})} = - \mathbb{E}_{(x,y)\sim\mathcal{D}} [\log{\sigma(\hat{r}(\rho_{i, c}) - \hat{r}(\rho_{i,r}))}] 
\end{equation}

In essence, during the RM training, we pass a preference pair $\{\rho_{i,c}, \rho_{i,r}\}$ labeled as $\rho_{i,c}$ winning over $\rho_{i,r}$ according to the annotator's preference $\mathbf{\Lambda}$. Provided sufficient preference samples in a dataset, the RM learns the winning preference features of the data that determine the winner in a pair, captured in the reward function $\hat{r}_{\mathbf{\Lambda}}$.
\begin{figure*}[t]
\centering
\begin{subfigure}{0.23\linewidth}
\centering
\vspace{-0.1in}
\includegraphics[width=\linewidth]{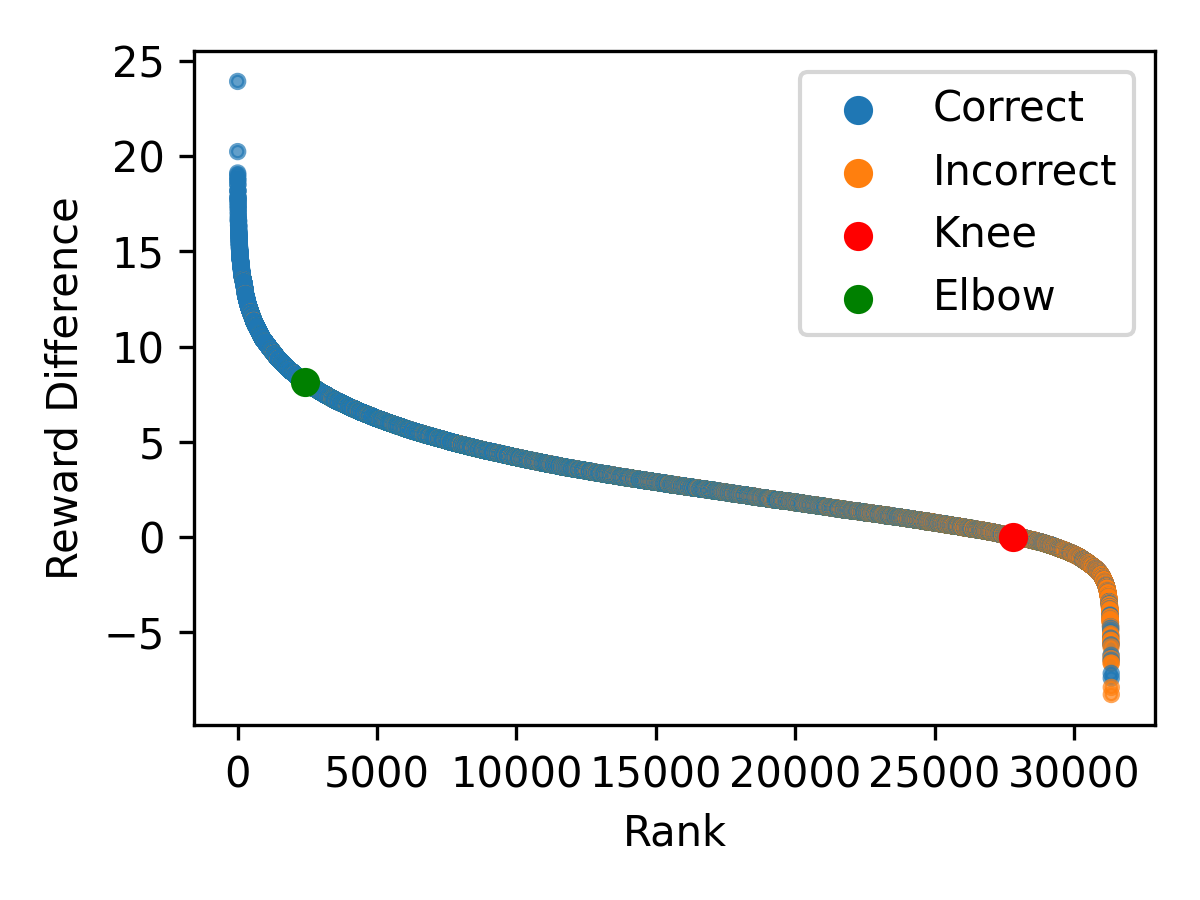}
\caption{Reward dist. : Itr-0}
\vspace{-0.1in}
\label{fig:hh_itr0_reward_curve}
\end{subfigure}
\begin{subfigure}{0.23\linewidth}
\centering
\includegraphics[width=\linewidth]{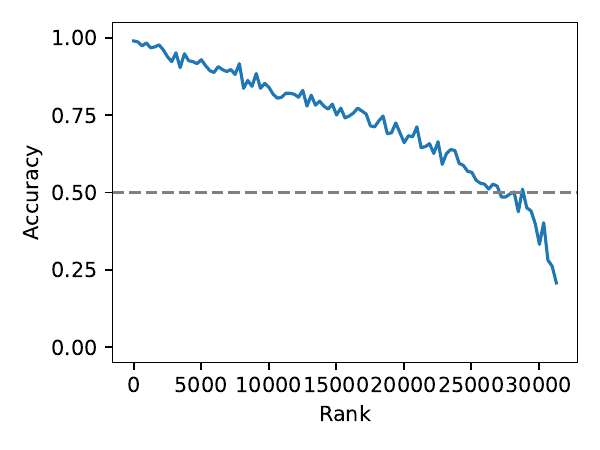}
\vspace{-0.1in}
\caption{Correctness dist. : Itr-0}
\vspace{-0.1in}
\label{fig:hh_itr0_accuracy_curve}
\end{subfigure}
\begin{subfigure}{0.23\linewidth}
\centering
\includegraphics[width=\linewidth]{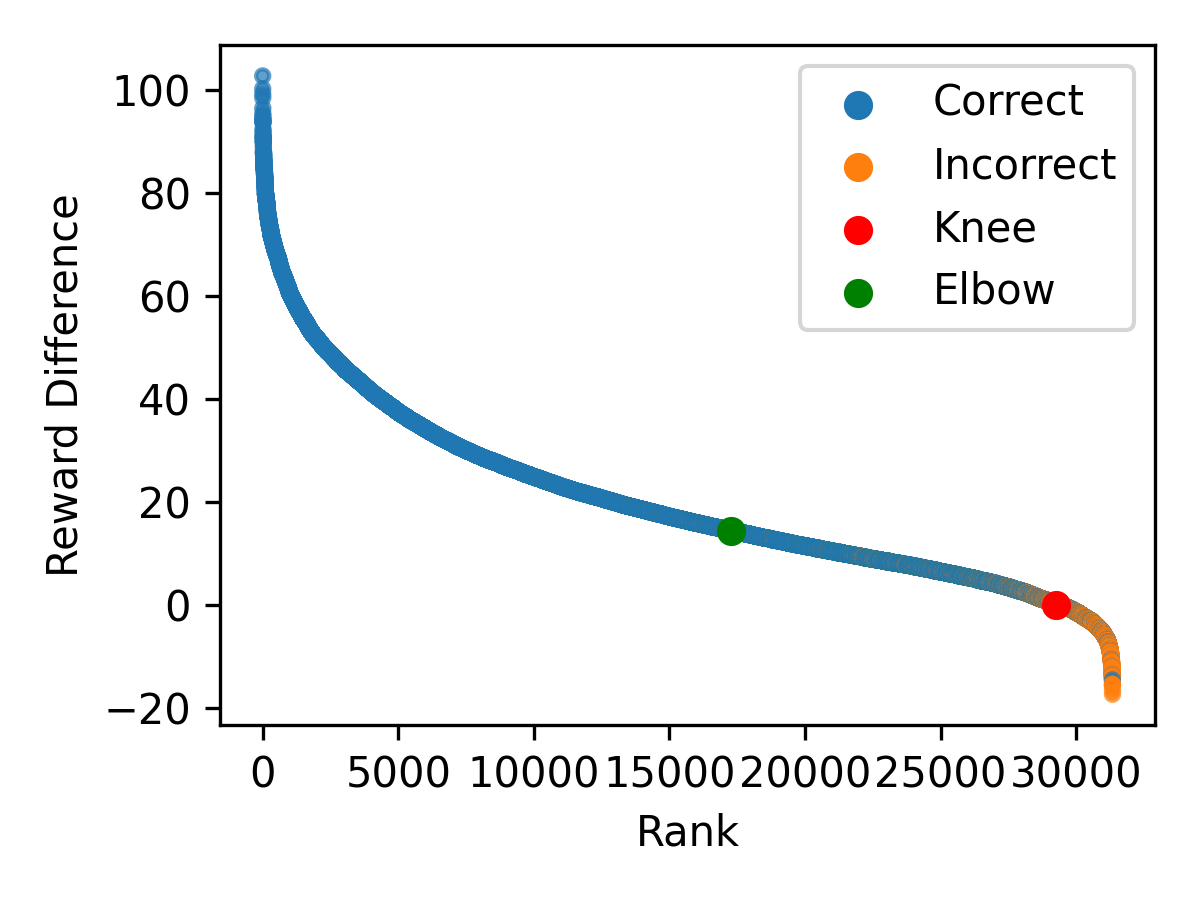}
\vspace{-0.1in}
\caption{Reward dist. : Itr-5}
\vspace{-0.1in}
\label{fig:hh_itr5_reward_curve}
\end{subfigure}
\begin{subfigure}{0.23\linewidth}
\centering
\includegraphics[width=\linewidth]{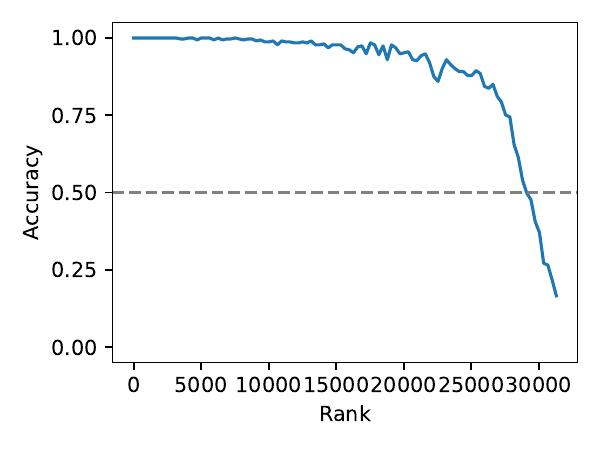}
\vspace{-0.1in}
\caption{Correctness dist. : Itr-5}
\vspace{-0.1in}
\label{fig:hh_itr5_accuracy_curve}
\end{subfigure}
\caption{Reward (assigned by a trained RM) and correctness (w.r.t. human preference) distribution curves for the very first and last iterations of \myname{}. These two types of curves provide the intuition of strategically selecting the samples for efficient human annotation towards improving alignment in the dataset. These curves further highlight the iterative refinement process, showing how alignment in the dataset progressively improves.}\vspace{-0.2in}
\label{fig:hh_reward_and_accuracy_curve}
\end{figure*}

\subsubsection{Looking at Reward Distribution}
\label{sec:leveraging_reward_score}
At this stage, we analyze the distribution of the predicted reward function ($\hat{r}_{\mathbf{\Lambda}}$) within the training preference dataset $\mathcal{D}_{\mathbf{\Lambda}}$. For each labeled preference pair $\{\rho_{i,c}, \rho_{i,r}\}$, we compute the reward score difference as $ \Delta_{\mathbf{\Lambda}}{\hat{r}_\mathbf{\Lambda}} = (\hat{r}_{\mathbf{\Lambda}}(\rho_{c}) - \hat{r}_{\mathbf{\Lambda}}(\rho_{r}))$. It is important to note that $\Delta_{\mathbf{\Lambda}}$ quantifies the relative preference score of a given pair in alignment with the annotator's preference orientation $\mathbf{\Lambda}$, satisfying the property $\Delta_{\mathbf{\Lambda}}{r} = - \Delta_{\mathbf{-\Lambda}}{r}$. By ranking all preference pairs in $\mathcal{D}_{\mathbf{\Lambda}}$ based on $\Delta_{\mathbf{\Lambda}}{\hat{r}_\mathbf{\Lambda}}$, a monotonic reward distribution curve, denoted as $\vartheta(\Delta_{\mathbf{\Lambda}}{\hat{r}_\mathbf{\Lambda})}$, emerges. This distribution, as depicted in Figure~\ref{fig:hh_itr0_reward_curve}, provides insight into the model’s reward assignment across the dataset, though for the moment, the legend in the graph can be disregarded. 

The reward distribution curve $\vartheta(\Delta_{\mathbf{\Lambda}}{\hat{r}_\mathbf{\Lambda})}$, derived from the training preference dataset $\mathcal{D}_{\mathbf{\Lambda}}$, reflects the degree of alignment the RM (trained with optimal validation loss) has achieved across $\mathcal{D}_{\mathbf{\Lambda}}$ during training. The upper left region of the curve consists of samples with high positive $\Delta_{\mathbf{\Lambda}}\hat{r}_\mathbf{\Lambda}$, indicating strong agreement between the RM and the training preference labels $\mathbf{\Lambda}$. This suggests that the RM effectively identifies and reinforces strong winning preference features in these samples, implying that these features were dominant in $\mathcal{D}_{\mathbf{\Lambda}}$. Conversely, the bottom right region of the curve contains samples with very low or even negative $\Delta_{\mathbf{\Lambda}}\hat{r}_\mathbf{\Lambda}$, signaling disagreement between the trained reward function $\hat{r}_\mathbf{\Lambda}$ and the training preference labels for these samples. This misalignment arises from two primary factors. (1) Absence of strong features, where RM is not able to find any strong preference feature in these samples according to $\hat{r}_\mathbf{\Lambda}$. (2) Conflicting samples within $\mathcal{D}_{\mathbf{\Lambda}}$, where the preference features of these samples are highly conflicting with other stronger preference features learned in $\hat{r}_\mathbf{\Lambda}$, leading the RM to penalize them.

\subsubsection{\myname{} Leveraging Reward Distribution}
\label{sec:leveraging_reward_score}
\myname{} trains the initial RM using a preference dataset labeled by a general-purpose LLM from the previous stage. We denote this dataset as $\mathcal{D}_{\mathbf{\Lambda_{LLM}}}$ where $\mathbf{\Lambda}_{LLM}$ represents the LLM's preference labeling. Since the RM training includes a validation set derived from $\mathcal{D}_{\mathbf{\Lambda_{LLM}}}$, this ensures that the trained RM is broadly aligned with the LLM’s preferences. We assume that the LLM has a coarse but reasonable understanding of preference judgments, particularly for relatively easy-to-annotate samples. As a result, the features of these samples dominate in $\hat{r}_\mathbf{\Lambda_{LLM}}$. Based on our earlier discussion, the upper left region of the reward density curve, $\vartheta(\Delta_{\mathbf{\Lambda_{LLM}}}{\hat{r}_\mathbf{\Lambda_{LLM}})}$ contains high density of samples with prominent preference features, i.e., those that are easier for the LLM to annotate accurately. Before proceeding, we further validate that the LLM is at least roughly aligned with the user in terms of these easy-to-annotate samples. This step mitigates the risk of significant misalignment due to prompt curation or model selection. To achieve this, \myname{} automatically (details in the following section) samples a small subset ($<0.1\%$) of preference data from the upper left region and gathers user feedback. If human agreement on these samples is low, it signals a major misalignment between the user and LLM. While we did not observe such cases in our experiments, this issue can be addressed by refining the judgment principles in the prompt. Updates can be directly made by the user, through incorporating verbose user feedback, or even through automated prompt optimization techniques~\cite{kepel2024autonomous}.

At this stage, we can identify regions with a high density of correctly labeled samples by the LLM, i.e., those that are relatively easy for the LLM to annotate in alignment with human preference. Now, we turn our attention to two critical types of samples necessary for achieving fine-grained alignment: (1) hard-to-annotate samples and (2) samples mislabeled by the LLM w.r.t. the human preference $\mathbf{\Lambda}_h$. Since the LLM was unable to correctly label these samples initially, the reward function $\hat{r}_\mathbf{\Lambda_{LLM}}$ cannot accurately capture their preference features. Consequently, these samples are expected to cluster around the bottom right region of the reward distribution curve $\vartheta(\Delta_{\mathbf{\Lambda_{LLM}}}{\hat{r}_\mathbf{\Lambda_{LLM}})}$. To illustrate this, we refer to Figure~\ref{fig:hh_itr0_reward_curve} and \ref{fig:hh_itr0_accuracy_curve}. Figure~\ref{fig:hh_itr0_reward_curve} shows $\vartheta(\Delta_{\mathbf{\Lambda_{LLM}}}{\hat{r}_\mathbf{\Lambda_{LLM}})}$ from one of our experiments. In this figure, we classify each sample $\rho_i \in \mathcal{D}_{\mathbf{\Lambda_{LLM}}}$ as either correctly or incorrectly labeled w.r.t. the human preference $\mathbf{\Lambda_{h}}$, i.e., whether the preference assigned by $\mathbf{\Lambda_{LLM}}$ is matching $\mathbf{\Lambda_{h}}$.
As observed, the upper left region of the curve contains a high density of correctly labeled samples, supporting our earlier claim that these represent the LLM’s easy-to-annotate cases. To quantify this, we generate an accuracy density curve for $\vartheta(\Delta_{\mathbf{\Lambda_{LLM}}}{\hat{r}_\mathbf{\Lambda_{LLM}})}$ w.r.t. the human preference $\mathbf{\Lambda_{h}}$, as shown in Figure~\ref{fig:hh_itr0_accuracy_curve}. This figure confirms that alignment with human preference decreases as we move towards the right side of the curve.   

While we can observe how alignment with $\mathbf{\Lambda_{h}}$ varies across the reward distribution curve, in real-world scenarios, we lack ground-truth labels to quantify this accuracy directly. Therefore, we need to \textit{estimate} the boundaries of key regions within the curve. To achieve this, we identify two strategic points: the ``elbow'' and the ``knee'', as illustrated in Figure~\ref{fig:hh_itr0_reward_curve}. These points correspond to sharp changes in $\Delta_{\mathbf{\Lambda_{LLM}}}{\hat{r}_\mathbf{\Lambda_{LLM}}}$, which we detect using the first-order derivative. The ``knee'' marks the transition to a region with lower accuracy density, whereas the ``elbow'' indicates a shift toward higher accuracy density. It is important to note that these points serve only as rough estimations of the region boundaries rather than precise demarcations.

\subsubsection{Selective Human Annotation}
To enhance alignment from this stage, human annotation is necessary, but it must be done efficiently to maximize its impact. A straightforward approach is referring to the accuracy density curve -- annotations in the lowest accuracy region would yield the highest benefit. Thus, we could start annotating from the very bottom of the curve. However, as previously discussed, some samples in this region may exhibit preference features that are largely opposite to the dominant features captured by $\hat{r}_\mathbf{\Lambda_{LLM}}$. These samples are highly likely to be mislabeled in $\mathbf{\Lambda_{LLM}}$ (see Appendix~\ref{appendix:iterative_improvement}). Instead of seeking human annotation, we can simply flip the preference of these samples to rectify. To estimate the location of such samples, we take the reflection of the ``elbow'' point w.r.t. the x-axis, as the elbow marks the region containing strong preference features. This ``reflection point'' always lies to the right of the ``knee'' in the lowest accuracy density region. We begin human annotation at this ``reflection point'' and proceed leftward along the curve, ensuring the most effective correction of alignment errors.

\subsubsection{Iterative Approach}
\label{sec:design:improve:iter}
The current reward function $\hat{r}_\mathbf{\Lambda_{LLM}}$, trained on $\mathcal{D}_{\mathbf{\Lambda_{LLM}}}$, exhibits an alignment gap w.r.t. $\mathbf{\Lambda_{h}}$ due to the presence of hard-to-annotate samples for the LLM and mislabeling by the LLM. Since we have identified ways to rectify these issues, we can refine $\mathcal{D}$ to improve alignment and train a new RM that better aligns with $\mathbf{\Lambda_{h}}$. Now, the question is how to prepare the dataset for the next iteration of RM training? Suppose we are currently in iteration 0 (Itr-0) with $\mathcal{D}_{\mathbf{\Lambda_{LLM}}}$ and $\hat{r}_\mathbf{\Lambda_{LLM}}$. For the iteration 1 (Itr-1) training dataset, $\mathcal{D}_{\mathbf{\Lambda_{Itr-1}^{T}}}^{T}$, our primary goal is to include high-confidence samples that are well-aligned with $\mathbf{\Lambda_{h}}$. The first choice is definitely human annotated samples from Itr-0.  Additionally, another set of candidates can be drawn from the high-accuracy density region of $\vartheta(\Delta_{\mathbf{\Lambda_{LLM}}}{\hat{r}_\mathbf{\Lambda_{LLM}})}$, specifically the region to the left of the ``elbow'', where the RM has learned strong preference features in alignment with $\mathbf{\Lambda_{h}}$.

Although these two sets of samples offer high precision, $\mathcal{D}_{\mathbf{\Lambda_{Itr-1}^{T}}}^{T}$ will still face a data coverage issue. Looking at the reward distribution curve, these two candidate sets represent samples with the longest distance, leaving gaps in middle region. However, expanding the dataset by including samples from the middle region, i.e., right of the ``elbow'' and left of the ``knee'' risks introducing misaligned samples. Since the accuracy in this region is likely to be just above $50\%$, obtaining human annotations for these samples would be inefficient. Furthermore, as the number of samples annotated from the right of the knee is relatively small, their preference features are likely to be overshadowed by the dominant preference features of the high numbers of left-side samples. As a result, their features may not be effectively captured in $\hat{r}_\mathbf{\Lambda_{Itr-1}^{T}}$. To balance these trade-offs, we introduce two hyperparameters, allowing for a more controlled and effective dataset expansion while maintaining alignment quality.

\begin{itemize}[leftmargin=*,topsep=2pt]
    \item \textbf{Back-off ratio ($\beta$)}: Determines how far to back off from the ``knee'' when selecting samples for the next iteration's dataset. A higher $\beta$ results in a more sanitized dataset, reducing noise but at the expense of lower data coverage.  
    \item \textbf{Amplification ratio ($\alpha$)}: Increases the influence of human-annotated samples by repeating them in the dataset, reinforcing their preference features in $\hat{r}_\mathbf{\Lambda_{Itr-1}^{T}}$. However, an excessively high $\alpha$ may lead to overfitting to selective human annotations.
\end{itemize}

The dataset $\mathcal{D}_{\mathbf{\Lambda_{Itr-1}}}^{T}$ consists of carefully selected samples from $\mathcal{D}_{\mathbf{\Lambda_{LLM}}}$, ensuring high alignment with $\mathbf{\Lambda_{h}}$ by optimally tuning the hyperparameters $\alpha$ and $\beta$. Training the RM on $\mathcal{D}_{\mathbf{\Lambda_{Itr-1}^{T}}}^{T}$ results in $\hat{r}_\mathbf{\Lambda_{Itr-1}^{T}}$, which is more closely aligned with $\mathbf{\Lambda_{h}}$. After training, we construct the dataset for generating the reward distribution curve by incorporating the remaining samples from Itr-0: $\mathcal{D}_{\mathbf{\Lambda_{Itr-1}}} = \mathcal{D}_{\mathbf{\Lambda_{Itr-1}^{T}}}^{T} \cup (\mathcal{D}_{\mathbf{\Lambda_{LLM}}} - \mathcal{D}_{\mathbf{\Lambda_{Itr-1}^{T}}}^{T})$. From this, we generate a new reward distribution curve, $\vartheta(\Delta_{\mathbf{\Lambda_{Itr-1}}}{\hat{r}_\mathbf{\Lambda_{Itr-1}^{T}})}$. While this curve demonstrates improved alignment with $\mathbf{\Lambda_{h}}$, full alignment is not necessarily achieved. However, it presents \myname{} with a distinct reward distribution curve compared to the previous iteration. This evolving diversity in $\vartheta(\cdot)$ enhances the variety of human annotations, maximizing the return on annotation investments and incrementally enriching $\mathcal{D}$. Note that the effectiveness of this diversification, as well as the corresponding improvements, depends on factors such as hyperparameter tuning (see Section~\ref{sec:results}), the original data distribution, and model selection.

\myname{} maximizes the efficiency of human annotations by iteratively refining $\vartheta(\cdot)$ and exposing annotators to diverse, LLM-mislabeled samples. To further enhance annotation efficiency, \myname{} employs random sharding to down-sample the original corpus. It begins by selecting a random shard of the dataset, iteratively improving alignment within that subset. Once the desired alignment is achieved, the final iteration's RM is used to label the entire corpus. This approach enables \myname{} to concentrate human annotations in a smaller, more targeted space while effectively propagating alignment across the full dataset at the end. 

\vspace{-0.1in}
\subsection{Reward Knowledge Transfer}
\vspace{-0.05in}

\myname{} progressively converges toward the oracle human preference through iterative RM training and strategic human annotation investment. As shown in Figure ~\ref{fig:hh_itr5_reward_curve} and ~\ref{fig:hh_itr5_accuracy_curve}, after five iterations, the reward distribution and accuracy curves closely align with the full-human annotation. Find the intermediate iteration curves in Appendix~\ref{appendix:iterative_improvement}. The required number of iterations depends on the available human annotation and RM training budget. Notably, full-human alignment can sometimes be achieved before exhausting the annotation budget. In such cases, the samples selected for human annotation would largely lack distinct preference features, indicating that the model has effectively captured the  human preference. Once desired alignment is achieved or the annotation budget is fully utilized, we proceed with fine-tuning an LLM for the downstream task. This can be done in two ways: 1) incorporating the final iteration RM into the PPO loop, or 2) labeling the whole dataset with the final RM and feeding the labeled dataset to a DPO pipeline.

\section{Results}
\label{sec:results}

In this section, we present the results of our main experiments, conducted on two datasets: HH-RLHF~\cite{bai2022training} and TL;DR~\cite{volske2017tl}. Specifically, we compare \myname{} against three baselines: (1) AI-only labeling, where samples are fully labeled by LLMs, (2) \textit{Random} human annotation, where samples are randomly selected for human annotation (matching the number of human-annotated samples in \myname{}), with the rest relying on AI feedback, and (2) \textit{Human}, where \textit{all} samples are annotated by humans. A detailed description of our experimental setup is provided in Appendix~\ref{appendix:setup}.
\vspace{-0.1in}

\subsection{Reward Modeling}

\subsubsection{Overall Alignment Improvement}
\label{sec:eval:rm:shard}

\begin{figure}[t]
    \centering
    \begin{subfigure}{0.92\linewidth}
        \centering        \includegraphics[width=\textwidth]{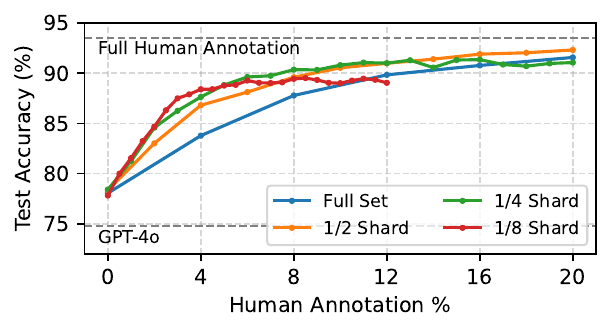}
        \vspace{-0.2in}
        \caption{HH-RLHF}
        \label{fig:hh_shard}
    \end{subfigure}
    
    \begin{subfigure}{0.92\linewidth}
        \centering
   \includegraphics[width=\textwidth]{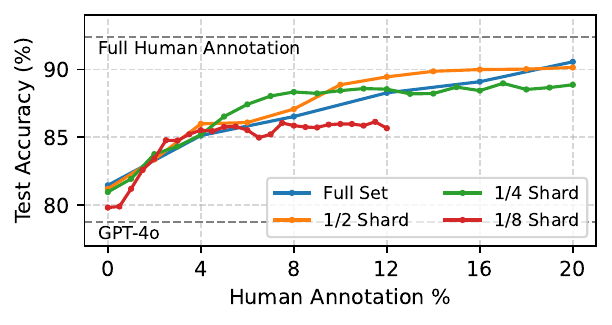}
    \vspace{-0.2in}
        \caption{TL;DR}
        \label{fig:tldr_shard}
    \end{subfigure}
    \vspace{-0.1in}
    \caption{Overall preference accuracy improvement of \myname{} on test data in an iterative manner. Here, we experiment with different sizes of down-sampled training data shards for \myname{}. We find 6\% (HH-RLHF) to 7\% (TL;DR) total human annotations on 1/4 shard yielding the optimal alignment for downstream tasks.}
    \label{fig:shard}
    \vspace{-0.2in}
\end{figure}

We here use GPT-4o for the initial alignment and evaluate \myname{}'s iterative alignment improvements by measuring the preference accuracy of RMs trained with varying proportions of human annotations relative to the full dataset. We employ \myname{} on both the complete dataset and multiple down-sampled shards as described in \S~\ref{sec:design:improve:iter}. For a given shard, we run \myname{} in an iterative manner infusing targeted human annotations in each iteration. We evaluate the trained RMs on a separate test dataset and report their preference accuracy in Figure~\ref{fig:shard}.

In Figure~\ref{fig:shard}, each data point for a shard corresponds to an iteration of \myname{}. The results show a consistent improvement in test preference accuracy across iterations, with significant early gains that gradually diminish as accuracy approaches the upper bound. Additionally, down-sampling enhances the efficiency of human annotations: \myname{} running on 1/2 and 1/4 shards outperforms its full-dataset counterpart when using the same number of human annotations. However, excessive down-sampling (e.g., 1/8 shard) may limit the achievable accuracy due to reduced data richness. For downstream task fine-tuning, we identify 1/4 shard as the optimal choice. Under this setting, \myname{} enhances preference accuracy on HH-RLHF from GPT-4o’s baseline of $74.7\%$ to $89.6\%$ with only $6\%$ human annotations, and on TL;DR, from $78.8\%$ to $88.0\%$ with just $7\%$ human annotations. We select the RMs from these iterations for labeling the full dataset, as outlined in \S~\ref{sec:design:improve:iter}. Evaluating the RMs trained on the fully human-labeled dataset, we observe the accuracy to be $91.8\%$ for HH-RLHF and $89.6\%$ for TL;DR.

% The results are shown in Figure~\ref{fig:shard}. When applied to a 1/4 shard (green), \myname{} improves preference accuracy on HH-RLHF from GPT-4o's $74.7\%$ to $89.6\%$ with only $6\%$ human annotations, and on TL;DR, from $78.8\%$ to $88.0\%$ with only $7\%$ human annotations.
% Notably, when we use these RMs to regenerate preferences on the full dataset and train a final RM, its test accuracy reaches $92.0??\%$ for HH-RLHF and $89.0??\%$ for TL;DR, which is less than $4??\%$ below full human annotation. Our experiments in \S~\ref{sec:eval:down} actually reveal that the remaining misaligned samples contain significant noise and do not meaningfully contribute to downstream tasks, and further alignment will lead to worse performance.

% Interestingly, sharding generally improves performance, as \myname{} operating on 1/2 and 1/4 shards tends to outperform its application on the full dataset with the same number of human annotations. However, excessively aggressive sharding (e.g., 1/8 shard) may limit the performance upper bound due to the lack of data richness and diversity.

\subsubsection{Comparison against the Baselines}
\label{sec:eval:rm:overall}
We begin by using two different LLMs -- GPT-4o and GPT-4o mini -- for the initial AI labeling. We then employ two separate \myname{} pipelines, \myname{} (4o) and \myname{} (4o mini), to improve alignment. To evaluate their effectiveness, we compare these pipelines against three aforementioned baselines (details in Appendix~\ref{appendix:setup:baselines}).

\begin{figure}
    \centering
    \includegraphics[width=0.99\linewidth]{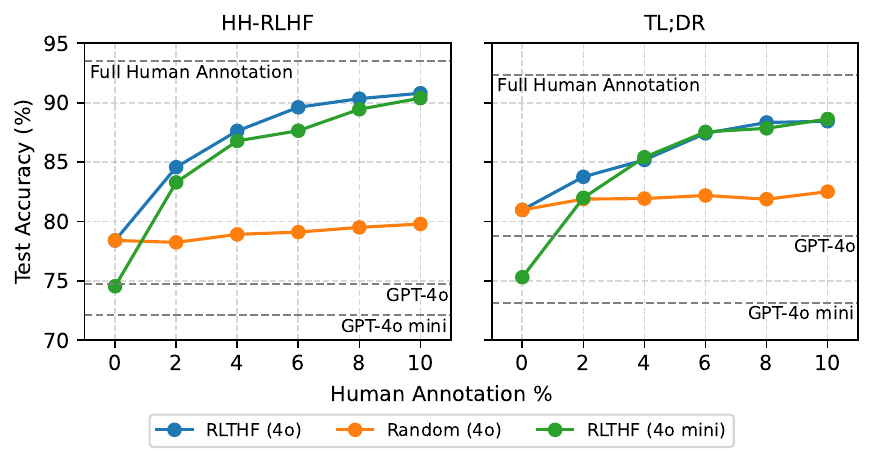}
    \vspace{-0.1in}
    \caption{\myname{} effectively bridges the prefernce accuracy gap between LLM-generated (GPT-4o/GPT-4o-mini) labels and fully human-annotated data, regardless of the initial labeling accuracy of the LLMs. By strategically incorporating human annotations, \myname{} achieves higher accuracy gains compared to random human annotation, maximizing the impact of human effort.}
    \label{fig:combined_iter_accuracy}
    \vspace{-0.2in}
\end{figure}

The results of this experimental setup on two datasets are shown in Figure~\ref{fig:combined_iter_accuracy}. \myname{} (4o) consistently outperforms \textit{Random} (4o), as random human annotation proves ineffective in correcting AI mislabeling, resulting in only marginal improvements in test accuracy. Of particular interest is the ``Return on Investment (ROI)'', which is measured as the increase in test accuracy per unit of human annotation. With just $6\%$ human annotation, \myname{} (4o) achieves a $15.9\times$ and $5.3\times$ higher ROI compared to \textit{Random} (4o) on HH-RLHF and TL;DR, respectively.

Notably, \myname{} remains robust even when the initial AI labeling quality is lower. While GPT-4o mini starts with an accuracy gap of $2.6\%$ and $5.6\%$ compared to GPT-4o, this gap shrinks to just $0.4\%$ and $-0.2\%$ after incorporating $10\%$ human annotation on HH-RLHF and TL;DR, respectively. This demonstrates that even when AI mislabeling is more prevalent, \myname{} more aggressively identifies and corrects errors, achieving a higher ROI on human annotation.

\subsubsection{Effects of Hyperparameters}
\label{sec:eval:rm:hyper}

\begin{figure*}[t]
    \centering
    % First row (HH-RLHF)
    \begin{subfigure}[b]{0.32\textwidth}
        \centering
        \includegraphics[width=\linewidth]{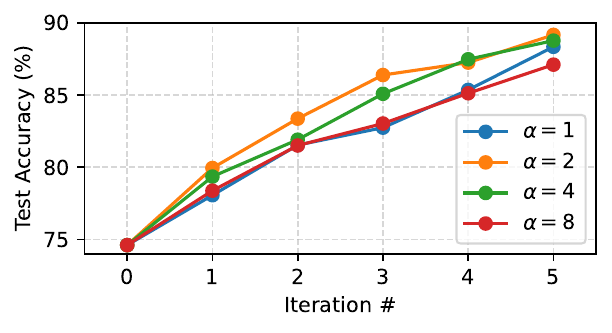}
        \caption{Amplification Ratio (HH-RLHF)}
        \label{fig:hh_amp}
    \end{subfigure}
    \hfill
    \begin{subfigure}[b]{0.32\textwidth}
        \centering
        \includegraphics[width=\linewidth]{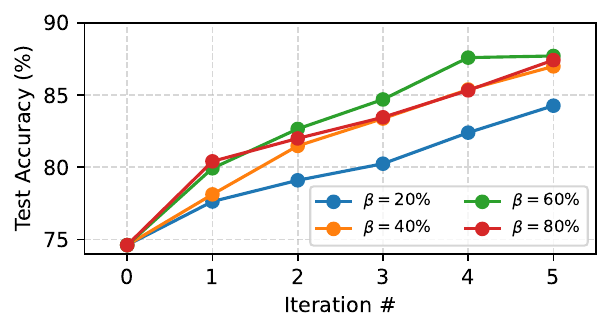}
        \caption{Back-off Ratio (HH-RLHF)}
        \label{fig:hh_backoff}
    \end{subfigure}
    \hfill
    \begin{subfigure}[b]{0.32\textwidth}
        \centering
        \includegraphics[width=\linewidth]{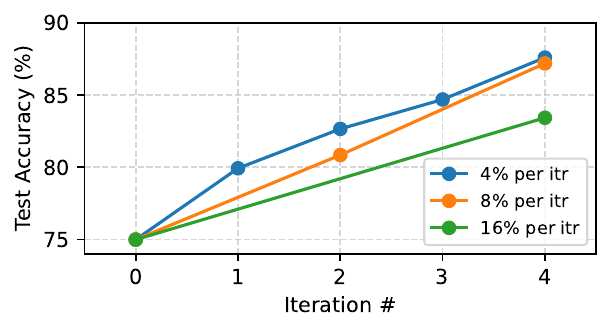}
        \caption{Annotation Batch Size (HH-RLHF)}
        \label{fig:hh_batch}
    \end{subfigure}

    % Second row (TL;DR)
    \begin{subfigure}[b]{0.32\textwidth}
        \centering
        \includegraphics[width=\linewidth]{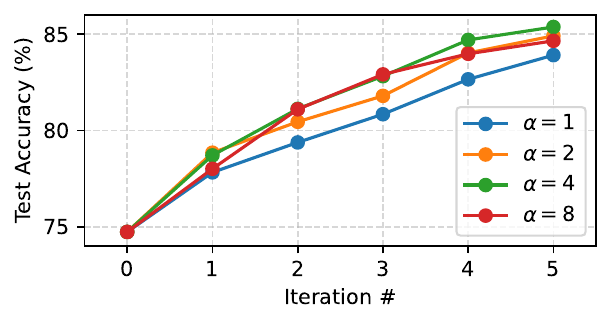}
        \caption{Amplification Ratio (TL;DR)}
        \label{fig:tldr_amp}
    \end{subfigure}
    \hfill
    \begin{subfigure}[b]{0.32\textwidth}
        \centering
        \includegraphics[width=\linewidth]{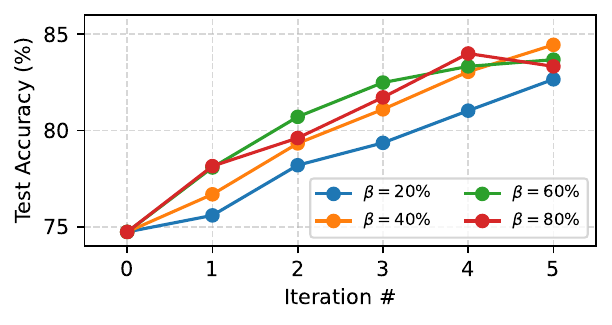}
        \caption{Back-off Ratio (TL;DR)}
        \label{fig:tldr_backoff}
    \end{subfigure}
    \hfill
    \begin{subfigure}[b]{0.32\textwidth}
        \centering
        \includegraphics[width=\linewidth]{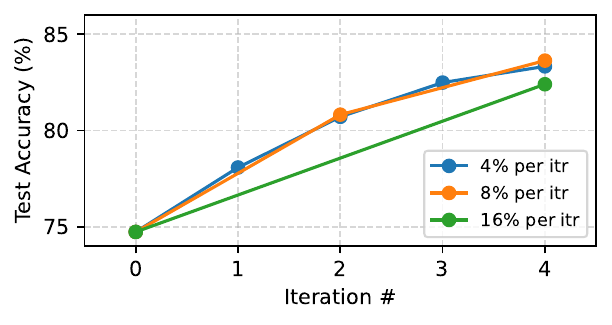}
        \caption{Annotation Batch Size (TL;DR)}
        \label{fig:tldr_batch}
    \end{subfigure}
    \vspace{-0.1in}
    \caption{Effects of \myname{} hyperparameters on HH-RLHF (top row) and TL;DR (bottom row). A larger amplification ratio ($\alpha$) and back-off ratio ($\beta$) are beneficial in the initial iterations but should be gradually reduced as human annotations accumulate and dataset gets sanitized. The human annotation budget should be distributed across multiple batches, though the impact of finer granularity remains unclear.}
    \label{fig:merged}
    \vspace{-0.1in}
\end{figure*}

\bbb{Amplification Ratio.}
To investigate how the amplification ratio (discussed in Section~\ref{sec:design:improve:iter}) $\alpha$ contributes to \myname{}, we fix the back-off ratio $\beta$ at $60\%$ and conduct a controlled study on different amplification ratios. The results for each dataset are shown in Figures~\ref{fig:hh_amp} and~\ref{fig:tldr_amp}. We observe that both no amplification ($\alpha=1$) and excessive amplification ($\alpha=8$) of human annotations lead to suboptimal RM improvements. Specifically, lower amplification results in smaller improvements in the initial iterations, while in later iterations, this trend reverses. This is expected, as no or low amplification weakens the impact of human annotations, particularly in the early iterations when the total number of annotations remains low, while over-amplification skews the training data distribution and increases the risk of overfitting, especially in later iterations when the base number of annotations is already large. For the best results, we start with a higher value of $\alpha$ and gradually reduce it (Appendix~\ref{appendix:setup:config}).

\begin{table*}[h]
    \centering
    \begin{tabular}{c|ccc|ccc}
        \toprule
        \multirow{2}{*}{Itr \#} & \multicolumn{3}{c|}{HH-RLHF} & \multicolumn{3}{c}{TL;DR} \\
        
         & No Annotation  & No Ampl./Back-off & \textbf{Full \myname{}}  & No Annotation  & No Ampl./Back-off  & \textbf{Full \myname{}}  \\
        \midrule
        0 & \multicolumn{3}{c|}{75.0} & \multicolumn{3}{c}{74.7} \\
        \midrule
        1 & 75.1 & 75.1 & \textbf{79.9} & 75.0 & 74.4 & \textbf{78.1} \\
        2 & 74.8 & 76.0 & \textbf{82.7} & 74.5 & 75.3 & \textbf{80.7} \\
        3 & 73.5 & 75.3 & \textbf{84.7} & 74.9 & 74.9 & \textbf{82.5} \\
        4 & 74.8 & 75.0 & \textbf{87.6} & 75.4 & 74.7 & \textbf{83.3} \\
        5 & 75.7 & 75.8 & \textbf{87.7} & 75.2 & 76.0 & \textbf{83.7} \\
        \bottomrule
    \end{tabular}
    \vspace{-0.1in}
    \caption{Ablation study of \myname{} on HH-RLHF and TL;DR. All three factors -- human annotation, amplification of human annotations, and the back-off mechanism -- play a crucial role in \myname{}'s effectiveness.}
    \label{tab:ablation}
    \vspace{-0.2in}
\end{table*}

\bbb{Back-off Ratio.}
To investigate the role of the back-off ratio $\beta$ \myname{}, we conduct a controlled study where all other hyperparameters are held at their default values. The results across datasets are presented in Figures~\ref{fig:hh_backoff} and~\ref{fig:tldr_backoff}. We observe a consistent pattern: a larger $\beta$ yields greater improvement during the initial iteration, but its benefit diminishes in later iterations. Conversely, a smaller $\beta$ leads to slower initial progress but accelerates in later stages. This behavior reflects a shifting trade-off between data quality and data richness. Early in training, when the dataset is relatively unsanitized, quality is the primary bottleneck—making a higher back-off ratio advantageous due to its stronger filtering effect. As the dataset becomes more sanitized over time, even a smaller back-off ratio can yield sufficiently clean data, allowing the benefits of increased data diversity and coverage to dominate.

In line with our configuration settings (Appendix~\ref{appendix:setup:config}), we recommend starting with a high back-off ratio and progressively reducing it as the dataset becomes more sanitized.

\bbb{Annotation Batch Size.}
To evaluate how the number of annotated samples per iteration impacts the effectiveness of \myname{} -- and more broadly, to assess whether \myname{} benefits from an iterative strategy over a one-shot annotate-all approach -- we conduct a controlled study, fixing the amplification ratio $\alpha$ at 4 and the back-off ratio $\beta$ at 60\%.

The results, shown in Figures~\ref{fig:hh_batch} and~\ref{fig:tldr_batch} (with annotation percentages computed relative to the shard rather than the full dataset), indicate that the iterative approach yields up to a $4.2\%$ improvement by Itr-4 compared to the one-shot annotation strategy. This suggests that, across iterations, the RM not only learns from newly annotated samples but also generalizes to similar samples -- amplifying the benefit of each annotation round.

However, our preliminary experiments show that further subdividing the annotation budget into smaller batches (e.g., 4\% vs. 8\% per iteration) does not provide a meaningful advantage. Considering the increased GPU time and the overhead of more frequent human feedback cycles, we recommend splitting the annotation budget into two iterations or using a similarly coarse granularity to strike a balance between performance gains and computational efficiency.

\subsubsection{Ablation Study}
\label{sec:eval:rm:ablation}
We conduct an ablation study to assess the necessity of two core mechanisms in \myname{} and understand their individual contributions. All experiments in this study are performed with $\alpha=4$, $\beta=60\%$, and $4\%$ human annotations per iteration, relative to a 1/4 shard of the dataset. The primary objective of this study is to address two key questions.

\bbb{Does self-improvement alone work?}
To evaluate whether \myname{} can operate effectively without human annotations, we set the annotation batch size to $0\%$, effectively reducing \myname{} to a purely self-improving framework, as proposed in recent studies~\cite{huang2024self, yu2025rip}. The results, shown in Table~\ref{tab:ablation}, reveal that self-improvement alone faces inherent limitations and fails to exceed the baseline preference accuracy achieved by the AI (GPT in this case). Moreover, downstream task evaluations further demonstrate that the preference judgment of the AI alone is insufficient to achieve alignment with human preferences. These findings indicate that self-improvement based solely on AI's preference is inadequate for oracle human alignment.

\bbb{Are amplification and back-off necessary?}
To assess the importance of \myname{}'s human annotation amplification and sanitization back-off mechanisms, we ablate both mechanisms simultaneously -- given that the effect of removing amplification alone was already examined in \S~\ref{sec:eval:rm:hyper} using $\alpha = 1$. As shown in Table~\ref{tab:ablation}, \myname{} achieves only marginal improvement under this configuration, even as human annotations accumulate over iterations. Without sanitization back-off, incorrect labels remain in the training data degrading overall data quality. Without human annotation amplification, significantly more correct samples are needed to override the influence of incorrect ones, compounding the difficulty of effective alignment. Together, these results highlight that both mechanisms are essential for driving meaningful improvements in \myname{}.

% \subsubsection{Density of correct labels along the curve (appendix)}
% \begin{itemize}
%     \item Case study on tldr
%     \item Case study on CUAD
% \end{itemize}

\subsection{Downstream Tasks}
\label{sec:eval:down}

To evaluate \myname{}'s effectiveness on final downstream tasks, we perform DPO training using data prepared by \myname{} and compare it against three baselines. For \myname{}, we use an RM trained on a 1/4 shard of the dataset -- refined with targeted human annotations -- to label preferences across the full corpus. Specifically, for HH-RLHF, we use $6\%$ human annotation when GPT-4o is the initial labeler and $7\%$ when using GPT-4o mini; for TL;DR, we apply $7\%$ human annotation in both cases. We then train an LLM, Qwen2.5-3B, with DPO using the \myname{}-refined dataset. For the \textit{Random} baseline, we inject an equivalent amount of randomly selected human annotations (on top of the AI-labeled data) into the DPO training set. Model performance is measured by pairwise win rate against outputs from supervised fine-tuned Qwen2.5-3B, using AlpacaEval~\cite{alpaca_eval} with Claude 3.5 Sonnet~\cite{anthropic_claude_sonnet} as the evaluation judge. Evaluations are conducted on held-out test sets from both datasets (details in Appendix~\ref{appendix:win_rate}).

\begin{table}[h]
    \centering
    \begin{tabular}{c|cc|cc}
        \toprule
        Dataset             & \multicolumn{2}{c|}{HH-RLHF} & \multicolumn{2}{c}{TL;DR} \\
        AI Labeler          & 4o & 4o-mini & 4o & 4o-mini \\
        \midrule
        \textit{AI-labeled} & 49.2 & 45.1 & 59.2 & 56.4 \\
        \textit{Random}     & 52.5 & 46.3 & 59.8 & 57.5 \\
        \textbf{\myname{}}  & \textbf{58.1} & \textbf{56.1} & \textbf{62.3} & \textbf{62.4} \\
        \midrule
        \textit{Human}      & \multicolumn{2}{c|}{55.7} & \multicolumn{2}{c}{60.2} \\
        \bottomrule
    \end{tabular}
    \vspace{-0.1in}
    \caption{Win rate against SFT (\%). \myname{} outperforms the \textit{AI-labeled}, \textit{Random}, and \textit{Human} baselines across both datasets. Even with weaker GPT-4o mini, \myname{} also achieves a win rate comparable to that with GPT-4o.}    
    \vspace{-0.2in}
    \label{tab:win_rate}
\end{table}

The results in Table~\ref{tab:win_rate} align with the observed preference accuracy trends. Across both datasets, \myname{} achieves a higher win rate than the fully \textit{Human}-annotated baseline using only $6$–$7\%$ of total human annotations, while significantly outperforming models trained on purely AI-labeled data as well as \textit{Random} human annotations. Notably, even with a weaker AI labeler (GPT-4o mini), \myname{} effectively bridges the gap within the same annotation budget, achieving a win rate comparable to that of GPT-4o. These findings are consistent with our observations in \S~\ref{sec:eval:rm:overall}, further validating \myname{}'s robustness and effectiveness, even when faced with suboptimal AI labeling due to model limitations, task complexity, or poor prompting.

Interestingly, \myname{} outperforms the fully \textit{Human}-annotated baseline, despite incorporating annotations from the same dataset. We attribute this advantage to \myname{}'s sanitized data selection for RM training, as discussed in \S~\ref{sec:design:improve:iter}. Fully human-annotated datasets inherently contain noise and biases~\cite{wang2024secrets, sun2024rethinking, ethayarajh2024kto}. In \S~\ref{sec:design:improve:iter}, we illustrated how such samples tend to cluster around the ``knee'' of the reward distribution curve. By leveraging the back-off ratio hyperparameter, \myname{} controls noisy and biased samples, ensuring a cleaner training dataset. The selected RMs from \myname{} are trained on data with a back-off ratio of $10\%$ in the corresponding iteration, resulting in reduced bias and noise. Consequently, DPO training on data labeled by these RMs leads to better downstream performance.

\subsection{Cost Analysis}
\myname{} introduces two types of additional costs: (1) the cost of LLM annotation during initial alignment, and (2) the cost of iterative RM training. However, our case study in Appendix~\ref{appendix:cost} demonstrates that even when accounting for these costs, \myname{} can still reduce the overall cost by 84.0--86.0\%, based on a conservative estimate.

\section{Conclusion}
 
In this work, we introduce \myname{}, an iterative reward model training approach that enhances alignment in preference datasets by strategically infusing human annotations, complemented by sanitized AI labeling. Through reward distribution analysis, we identify key samples for targeted human intervention, optimizing annotation efficiency. Our experiments demonstrate that \myname{} progressively improves alignment, converging toward comprehensive human alignment. Furthermore, models trained on our refined datasets for downstream tasks even outperform the models trained on datasets with full-human annotations.

% \section*{Acknowledgments}
% We thank Eduardo Rodrigues, Roberto Estevao, Maria Angels de Luis Balaguer, Jessica Wolk, Rafael Padilha, Leonardo Nunes, and Shobana Balakrishnan for their valuable support in experimenting with private data and integrating RLTHF into the production pipeline.

% \textbf{Do not} include acknowledgements in the initial version of
% the paper submitted for blind review.

% If a paper is accepted, the final camera-ready version can (and
% usually should) include acknowledgements.  Such acknowledgements
% should be placed at the end of the section, in an unnumbered section
% that does not count towards the paper page limit. Typically, this will 
% include thanks to reviewers who gave useful comments, to colleagues 
% who contributed to the ideas, and to funding agencies and corporate 
% sponsors that provided financial support.

\section*{Impact Statement}

This paper presents work whose goal is to advance the field of 
Machine Learning. There are many potential societal consequences 
of our work, none which we feel must be specifically highlighted here.

% % In the unusual situation where you want a paper to appear in the
% % references without citing it in the main text, use \nocite

\bibliography{reference}

\begin{thebibliography}{47}
\providecommand{\natexlab}[1]{#1}
\providecommand{\url}[1]{\texttt{#1}}
\expandafter\ifx\csname urlstyle\endcsname\relax
  \providecommand{\doi}[1]{doi: #1}\else
  \providecommand{\doi}{doi: \begingroup \urlstyle{rm}\Url}\fi

\bibitem[Achiam et~al.(2023)Achiam, Adler, Agarwal, Ahmad, Akkaya, Aleman,
  Almeida, Altenschmidt, Altman, Anadkat, et~al.]{achiam2023gpt}
Achiam, J., Adler, S., Agarwal, S., Ahmad, L., Akkaya, I., Aleman, F.~L.,
  Almeida, D., Altenschmidt, J., Altman, S., Anadkat, S., et~al.
\newblock Gpt-4 technical report.
\newblock \emph{arXiv preprint arXiv:2303.08774}, 2023.

\bibitem[AI(2024)]{folio}
AI, F.
\newblock {LLM FINE-TUNING SERVICES BY FOLIO3 AI}, 2024.

\bibitem[{Amazon Web Services}(2025)]{aws_groundtruth_pricing}
{Amazon Web Services}.
\newblock Amazon sagemaker ground truth pricing, 2025.
\newblock URL \url{https://aws.amazon.com/sagemaker-ai/groundtruth/pricing/}.
\newblock Accessed: 2025-03-31.

\bibitem[{Anthropic}(2024)]{anthropic_claude_sonnet}
{Anthropic}.
\newblock {Claude 3.5 Sonnet}.
\newblock \url{https://www.anthropic.com/claude/sonnet}, 2024.
\newblock Accessed: 2025-02-18.

\bibitem[Atreya(2024)]{ft_service}
Atreya, M.
\newblock {Fine-Tuning AI Models with Tuning-as-a-Service Platforms}, 2024.

\bibitem[Bai et~al.(2022{\natexlab{a}})Bai, Jones, Ndousse, Askell, Chen,
  DasSarma, Drain, Fort, Ganguli, Henighan, et~al.]{bai2022training}
Bai, Y., Jones, A., Ndousse, K., Askell, A., Chen, A., DasSarma, N., Drain, D.,
  Fort, S., Ganguli, D., Henighan, T., et~al.
\newblock Training a helpful and harmless assistant with reinforcement learning
  from human feedback.
\newblock \emph{arXiv preprint arXiv:2204.05862}, 2022{\natexlab{a}}.

\bibitem[Bai et~al.(2022{\natexlab{b}})Bai, Kadavath, Kundu, Askell, Kernion,
  Jones, Chen, Goldie, Mirhoseini, McKinnon, et~al.]{bai2022constitutional}
Bai, Y., Kadavath, S., Kundu, S., Askell, A., Kernion, J., Jones, A., Chen, A.,
  Goldie, A., Mirhoseini, A., McKinnon, C., et~al.
\newblock Constitutional ai: Harmlessness from ai feedback.
\newblock \emph{arXiv preprint arXiv:2212.08073}, 2022{\natexlab{b}}.

\bibitem[Chung et~al.(2024)Chung, Hou, Longpre, Zoph, Tay, Fedus, Li, Wang,
  Dehghani, Brahma, et~al.]{chung2024scaling}
Chung, H.~W., Hou, L., Longpre, S., Zoph, B., Tay, Y., Fedus, W., Li, Y., Wang,
  X., Dehghani, M., Brahma, S., et~al.
\newblock Scaling instruction-finetuned language models.
\newblock \emph{Journal of Machine Learning Research}, 25\penalty0
  (70):\penalty0 1--53, 2024.

\bibitem[David(1963)]{david1963method}
David, H.~A.
\newblock \emph{The method of paired comparisons}, volume~12.
\newblock London, 1963.

\bibitem[Dubey et~al.(2024)Dubey, Jauhri, Pandey, Kadian, Al-Dahle, Letman,
  Mathur, Schelten, Yang, Fan, et~al.]{dubey2024llama}
Dubey, A., Jauhri, A., Pandey, A., Kadian, A., Al-Dahle, A., Letman, A.,
  Mathur, A., Schelten, A., Yang, A., Fan, A., et~al.
\newblock The llama 3 herd of models.
\newblock \emph{arXiv preprint arXiv:2407.21783}, 2024.

\bibitem[Ethayarajh et~al.(2024)Ethayarajh, Xu, Muennighoff, Jurafsky, and
  Kiela]{ethayarajh2024kto}
Ethayarajh, K., Xu, W., Muennighoff, N., Jurafsky, D., and Kiela, D.
\newblock Kto: Model alignment as prospect theoretic optimization.
\newblock \emph{arXiv preprint arXiv:2402.01306}, 2024.

\bibitem[Huang et~al.(2024)Huang, Fan, Wang, Yang, Zhao, Lin, Lin, Zhang,
  Rajmohan, and Zhang]{huang2024self}
Huang, C., Fan, Z., Wang, L., Yang, F., Zhao, P., Lin, Z., Lin, Q., Zhang, D.,
  Rajmohan, S., and Zhang, Q.
\newblock Self-evolved reward learning for llms.
\newblock \emph{arXiv preprint arXiv:2411.00418}, 2024.

\bibitem[Huang et~al.(2023)Huang, Chen, Mishra, Zheng, Yu, Song, and
  Zhou]{huang2023large}
Huang, J., Chen, X., Mishra, S., Zheng, H.~S., Yu, A.~W., Song, X., and Zhou,
  D.
\newblock Large language models cannot self-correct reasoning yet.
\newblock \emph{arXiv preprint arXiv:2310.01798}, 2023.

\bibitem[Jiang et~al.(2024)Jiang, Sablayrolles, Roux, Mensch, Savary, Bamford,
  Chaplot, Casas, Hanna, Bressand, et~al.]{jiang2024mixtral}
Jiang, A.~Q., Sablayrolles, A., Roux, A., Mensch, A., Savary, B., Bamford, C.,
  Chaplot, D.~S., Casas, D. d.~l., Hanna, E.~B., Bressand, F., et~al.
\newblock Mixtral of experts.
\newblock \emph{arXiv preprint arXiv:2401.04088}, 2024.

\bibitem[Kepel \& Valogianni(2024)Kepel and Valogianni]{kepel2024autonomous}
Kepel, D. and Valogianni, K.
\newblock Autonomous prompt engineering in large language models.
\newblock \emph{arXiv preprint arXiv:2407.11000}, 2024.

\bibitem[K{\"o}pf et~al.(2024)K{\"o}pf, Kilcher, von R{\"u}tte, Anagnostidis,
  Tam, Stevens, Barhoum, Nguyen, Stanley, Nagyfi,
  et~al.]{kopf2024openassistant}
K{\"o}pf, A., Kilcher, Y., von R{\"u}tte, D., Anagnostidis, S., Tam, Z.~R.,
  Stevens, K., Barhoum, A., Nguyen, D., Stanley, O., Nagyfi, R., et~al.
\newblock Openassistant conversations-democratizing large language model
  alignment.
\newblock \emph{Advances in Neural Information Processing Systems}, 36, 2024.

\bibitem[Lee et~al.()Lee, Phatale, Mansoor, Mesnard, Ferret, Lu, Bishop, Hall,
  Carbune, Rastogi, et~al.]{leerlaif}
Lee, H., Phatale, S., Mansoor, H., Mesnard, T., Ferret, J., Lu, K.~R., Bishop,
  C., Hall, E., Carbune, V., Rastogi, A., et~al.
\newblock Rlaif vs. rlhf: Scaling reinforcement learning from human feedback
  with ai feedback.
\newblock In \emph{Forty-first International Conference on Machine Learning}.

\bibitem[Lee et~al.(2023)Lee, Phatale, Mansoor, Lu, Mesnard, Ferret, Bishop,
  Hall, Carbune, and Rastogi]{lee2023rlaif}
Lee, H., Phatale, S., Mansoor, H., Lu, K.~R., Mesnard, T., Ferret, J., Bishop,
  C., Hall, E., Carbune, V., and Rastogi, A.
\newblock Rlaif: Scaling reinforcement learning from human feedback with ai
  feedback.
\newblock 2023.

\bibitem[Li et~al.(2023{\natexlab{a}})Li, Yu, Zhou, Schick, Levy, Zettlemoyer,
  Weston, and Lewis]{li2023self}
Li, X., Yu, P., Zhou, C., Schick, T., Levy, O., Zettlemoyer, L., Weston, J.,
  and Lewis, M.
\newblock Self-alignment with instruction backtranslation.
\newblock \emph{arXiv preprint arXiv:2308.06259}, 2023{\natexlab{a}}.

\bibitem[Li et~al.(2023{\natexlab{b}})Li, Zhang, Dubois, Taori, Gulrajani,
  Guestrin, Liang, and Hashimoto]{alpaca_eval}
Li, X., Zhang, T., Dubois, Y., Taori, R., Gulrajani, I., Guestrin, C., Liang,
  P., and Hashimoto, T.~B.
\newblock Alpacaeval: An automatic evaluator of instruction-following models.
\newblock \url{https://github.com/tatsu-lab/alpaca_eval}, 5 2023{\natexlab{b}}.

\bibitem[Microsoft(2024)]{m365}
Microsoft.
\newblock {Microsoft 365 Copilot Tuning overview (preview)}, 2024.
\newblock URL
  \url{https://learn.microsoft.com/en-us/copilot/microsoft-365/copilot-tuning-overview}.

\bibitem[{Microsoft Azure}(2025)]{azure_ndma100v4}
{Microsoft Azure}.
\newblock Ndm a100 v4-series virtual machines, 2025.
\newblock URL
  \url{https://learn.microsoft.com/en-us/azure/virtual-machines/sizes/gpu-accelerated/ndma100v4-series?tabs=sizebasic}.
\newblock Accessed: 2025-03-31.

\bibitem[{OpenAI}(2025)]{openai_api_pricing}
{OpenAI}.
\newblock Openai api pricing, 2025.
\newblock URL \url{https://openai.com/api/pricing/}.
\newblock Accessed: 2025-03-31.

\bibitem[Ouyang et~al.(2022)Ouyang, Wu, Jiang, Almeida, Wainwright, Mishkin,
  Zhang, Agarwal, Slama, Ray, et~al.]{ouyang2022training}
Ouyang, L., Wu, J., Jiang, X., Almeida, D., Wainwright, C., Mishkin, P., Zhang,
  C., Agarwal, S., Slama, K., Ray, A., et~al.
\newblock Training language models to follow instructions with human feedback.
\newblock \emph{Advances in neural information processing systems},
  35:\penalty0 27730--27744, 2022.

\bibitem[Pang et~al.(2023)Pang, Wang, Li, Chen, Xu, Zhang, and
  Yu]{pang2023language}
Pang, J.-C., Wang, P., Li, K., Chen, X.-H., Xu, J., Zhang, Z., and Yu, Y.
\newblock Language model self-improvement by reinforcement learning
  contemplation.
\newblock \emph{arXiv preprint arXiv:2305.14483}, 2023.

\bibitem[Panickssery et~al.(2024)Panickssery, Bowman, and
  Feng]{panickssery2024llm}
Panickssery, A., Bowman, S.~R., and Feng, S.
\newblock Llm evaluators recognize and favor their own generations.
\newblock \emph{arXiv preprint arXiv:2404.13076}, 2024.

\bibitem[Rafailov et~al.(2024)Rafailov, Sharma, Mitchell, Manning, Ermon, and
  Finn]{rafailov2024direct}
Rafailov, R., Sharma, A., Mitchell, E., Manning, C.~D., Ermon, S., and Finn, C.
\newblock Direct preference optimization: Your language model is secretly a
  reward model.
\newblock \emph{Advances in Neural Information Processing Systems}, 36, 2024.

\bibitem[Schulman et~al.(2017)Schulman, Wolski, Dhariwal, Radford, and
  Klimov]{schulman2017proximal}
Schulman, J., Wolski, F., Dhariwal, P., Radford, A., and Klimov, O.
\newblock Proximal policy optimization algorithms.
\newblock \emph{arXiv preprint arXiv:1707.06347}, 2017.

\bibitem[Sharma(2024)]{nuances}
Sharma, A.
\newblock {Announcing fine-tuning for customization and support for new models
  in Azure AI}, 2024.

\bibitem[Sharma et~al.(2024)Sharma, Keh, Mitchell, Finn, Arora, and
  Kollar]{sharma2024critical}
Sharma, A., Keh, S., Mitchell, E., Finn, C., Arora, K., and Kollar, T.
\newblock A critical evaluation of ai feedback for aligning large language
  models.
\newblock \emph{arXiv preprint arXiv:2402.12366}, 2024.

\bibitem[Snell et~al.(2024)Snell, Lee, Xu, and Kumar]{snell2024scaling}
Snell, C., Lee, J., Xu, K., and Kumar, A.
\newblock Scaling llm test-time compute optimally can be more effective than
  scaling model parameters.
\newblock \emph{arXiv preprint arXiv:2408.03314}, 2024.

\bibitem[Stiennon et~al.(2020)Stiennon, Ouyang, Wu, Ziegler, Lowe, Voss,
  Radford, Amodei, and Christiano]{stiennon2020learning}
Stiennon, N., Ouyang, L., Wu, J., Ziegler, D., Lowe, R., Voss, C., Radford, A.,
  Amodei, D., and Christiano, P.~F.
\newblock Learning to summarize with human feedback.
\newblock \emph{Advances in Neural Information Processing Systems},
  33:\penalty0 3008--3021, 2020.

\bibitem[Sun et~al.(2024{\natexlab{a}})Sun, Shen, and Ton]{sun2024rethinking}
Sun, H., Shen, Y., and Ton, J.-F.
\newblock Rethinking bradley-terry models in preference-based reward modeling:
  Foundations, theory, and alternatives.
\newblock \emph{arXiv preprint arXiv:2411.04991}, 2024{\natexlab{a}}.

\bibitem[Sun et~al.(2024{\natexlab{b}})Sun, Shen, Zhou, Zhang, Chen, Cox, Yang,
  and Gan]{sun2024principle}
Sun, Z., Shen, Y., Zhou, Q., Zhang, H., Chen, Z., Cox, D., Yang, Y., and Gan,
  C.
\newblock Principle-driven self-alignment of language models from scratch with
  minimal human supervision.
\newblock \emph{Advances in Neural Information Processing Systems}, 36,
  2024{\natexlab{b}}.

\bibitem[Team et~al.(2023)Team, Anil, Borgeaud, Alayrac, Yu, Soricut,
  Schalkwyk, Dai, Hauth, Millican, et~al.]{team2023gemini}
Team, G., Anil, R., Borgeaud, S., Alayrac, J.-B., Yu, J., Soricut, R.,
  Schalkwyk, J., Dai, A.~M., Hauth, A., Millican, K., et~al.
\newblock Gemini: a family of highly capable multimodal models.
\newblock \emph{arXiv preprint arXiv:2312.11805}, 2023.

\bibitem[Thoppilan et~al.(2022)Thoppilan, De~Freitas, Hall, Shazeer,
  Kulshreshtha, Cheng, Jin, Bos, Baker, Du, et~al.]{thoppilan2022lamda}
Thoppilan, R., De~Freitas, D., Hall, J., Shazeer, N., Kulshreshtha, A., Cheng,
  H.-T., Jin, A., Bos, T., Baker, L., Du, Y., et~al.
\newblock Lamda: Language models for dialog applications.
\newblock \emph{arXiv preprint arXiv:2201.08239}, 2022.

\bibitem[Touvron et~al.(2023)Touvron, Lavril, Izacard, Martinet, Lachaux,
  Lacroix, Rozi{\`e}re, Goyal, Hambro, Azhar, et~al.]{touvron2023llama}
Touvron, H., Lavril, T., Izacard, G., Martinet, X., Lachaux, M.-A., Lacroix,
  T., Rozi{\`e}re, B., Goyal, N., Hambro, E., Azhar, F., et~al.
\newblock Llama: Open and efficient foundation language models.
\newblock \emph{arXiv preprint arXiv:2302.13971}, 2023.

\bibitem[V{\"o}lske et~al.(2017)V{\"o}lske, Potthast, Syed, and
  Stein]{volske2017tl}
V{\"o}lske, M., Potthast, M., Syed, S., and Stein, B.
\newblock Tl; dr: Mining reddit to learn automatic summarization.
\newblock In \emph{Proceedings of the Workshop on New Frontiers in
  Summarization}, pp.\  59--63, 2017.

\bibitem[Wang et~al.(2024{\natexlab{a}})Wang, Zheng, Chen, Liu, Dou, Huang,
  Shen, Jin, Zhou, Shi, et~al.]{wang2024secrets}
Wang, B., Zheng, R., Chen, L., Liu, Y., Dou, S., Huang, C., Shen, W., Jin, S.,
  Zhou, E., Shi, C., et~al.
\newblock Secrets of rlhf in large language models part ii: Reward modeling.
\newblock \emph{arXiv preprint arXiv:2401.06080}, 2024{\natexlab{a}}.

\bibitem[Wang et~al.(2024{\natexlab{b}})Wang, Li, Shao, Xu, Dai, Li, Chen, Wu,
  and Sui]{wang2024math}
Wang, P., Li, L., Shao, Z., Xu, R., Dai, D., Li, Y., Chen, D., Wu, Y., and Sui,
  Z.
\newblock Math-shepherd: Verify and reinforce llms step-by-step without human
  annotations.
\newblock In \emph{Proceedings of the 62nd Annual Meeting of the Association
  for Computational Linguistics (Volume 1: Long Papers)}, pp.\  9426--9439,
  2024{\natexlab{b}}.

\bibitem[Wei et~al.(2021)Wei, Bosma, Zhao, Guu, Yu, Lester, Du, Dai, and
  Le]{wei2021finetuned}
Wei, J., Bosma, M., Zhao, V.~Y., Guu, K., Yu, A.~W., Lester, B., Du, N., Dai,
  A.~M., and Le, Q.~V.
\newblock Finetuned language models are zero-shot learners.
\newblock \emph{arXiv preprint arXiv:2109.01652}, 2021.

\bibitem[Xu et~al.(2024)Xu, Fu, Gao, Ye, Liu, Mei, Wang, Yu, and Wu]{xu2024dpo}
Xu, S., Fu, W., Gao, J., Ye, W., Liu, W., Mei, Z., Wang, G., Yu, C., and Wu, Y.
\newblock Is dpo superior to ppo for llm alignment? a comprehensive study.
\newblock \emph{arXiv preprint arXiv:2404.10719}, 2024.

\bibitem[Yu et~al.(2025)Yu, Yuan, Golovneva, Wu, Sukhbaatar, Weston, and
  Xu]{yu2025rip}
Yu, P., Yuan, W., Golovneva, O., Wu, T., Sukhbaatar, S., Weston, J., and Xu, J.
\newblock Rip: Better models by survival of the fittest prompts.
\newblock \emph{arXiv preprint arXiv:2501.18578}, 2025.

\bibitem[Yuan et~al.(2024)Yuan, Pang, Cho, Sukhbaatar, Xu, and
  Weston]{yuan2024self}
Yuan, W., Pang, R.~Y., Cho, K., Sukhbaatar, S., Xu, J., and Weston, J.
\newblock Self-rewarding language models.
\newblock \emph{arXiv preprint arXiv:2401.10020}, 2024.

\bibitem[Zeng et~al.(2024)Zeng, Cheng, Yin, Wang, Li, Zhou, Guo, Huang, and
  Qiu]{zeng2024scaling}
Zeng, Z., Cheng, Q., Yin, Z., Wang, B., Li, S., Zhou, Y., Guo, Q., Huang, X.,
  and Qiu, X.
\newblock Scaling of search and learning: A roadmap to reproduce o1 from
  reinforcement learning perspective.
\newblock \emph{arXiv preprint arXiv:2412.14135}, 2024.

\bibitem[Zheng et~al.(2023{\natexlab{a}})Zheng, Chiang, Sheng, Zhuang, Wu,
  Zhuang, Lin, Li, Li, Xing, et~al.]{zheng2023judging}
Zheng, L., Chiang, W.-L., Sheng, Y., Zhuang, S., Wu, Z., Zhuang, Y., Lin, Z.,
  Li, Z., Li, D., Xing, E., et~al.
\newblock Judging llm-as-a-judge with mt-bench and chatbot arena.
\newblock \emph{Advances in Neural Information Processing Systems},
  36:\penalty0 46595--46623, 2023{\natexlab{a}}.

\bibitem[Zheng et~al.(2023{\natexlab{b}})Zheng, Dou, Gao, Hua, Shen, Wang, Liu,
  Jin, Liu, Zhou, et~al.]{zheng2023secrets}
Zheng, R., Dou, S., Gao, S., Hua, Y., Shen, W., Wang, B., Liu, Y., Jin, S.,
  Liu, Q., Zhou, Y., et~al.
\newblock Secrets of rlhf in large language models part i: Ppo.
\newblock \emph{arXiv preprint arXiv:2307.04964}, 2023{\natexlab{b}}.

\end{thebibliography}
\bibliographystyle{icml2025}

% %%%%%%%%%%%%%%%%%%%%%%%%%%%%%%%%%%%%%%%%%%%%%%%%%%%%%%%%%%%%%%%%%%%%%%%%%%%%%%%
% %%%%%%%%%%%%%%%%%%%%%%%%%%%%%%%%%%%%%%%%%%%%%%%%%%%%%%%%%%%%%%%%%%%%%%%%%%%%%%%
% % APPENDIX
% %%%%%%%%%%%%%%%%%%%%%%%%%%%%%%%%%%%%%%%%%%%%%%%%%%%%%%%%%%%%%%%%%%%%%%%%%%%%%%%
% %%%%%%%%%%%%%%%%%%%%%%%%%%%%%%%%%%%%%%%%%%%%%%%%%%%%%%%%%%%%%%%%%%%%%%%%%%%%%%%
\newpage
\appendix
\onecolumn
\section{Prompt Templates for Initial Alignment}
\label{appendix:prompt}
We structure the prompt using the following components to obtain the initial preference judgments from off-the-shelf LLMs (e.g., GPT-4o, Gemini, etc.).  
\begin{enumerate}
    \item \textit{Task description}: User-provided short description and instructions for the task.
    \item \textit{Preference criteria}: User-provided detailed criteria defining what is a good/bad response.
    \item \textit{Few-shot examples (optional)}: User-provided samples, chain-of-thought rationales and final preferences.
    \item \textit{Sample to annotate}: The target sample to be annotated.
\end{enumerate}
Below we provide our unified system prompt and user prompts used for HH-RLHF and TL;DR. Note that these prompts are not carefully tuned, as \myname{} should not assume ideal prompt engineering.

\subsection{System Prompt}
\fbox{
\begin{minipage}{\textwidth}
\ttfamily

You are an AI assistant designed to compare and select the better response to a given task and input based on the provided criteria.

\end{minipage}
}

\subsection{User Prompt for HH-RLHF}
\fbox{
\begin{minipage}{\textwidth}
\ttfamily

\#\# TASK
\\
You are a helpful and harmless Assistant. Given the context, provide the most helpful and harmless response.

-------------------------------------------------

\#\# CRITERIA

- **Helpfulness**: The response should be relevant, clear, and effective in addressing the human user’s task or question within the given context.

- **Honesty**: The response should be accurate, truthful, and reliable.

- **Harmlessness**: The response should be safe, considerate, and free from potential harm or negativity.

-------------------------------------------------

\#\# INPUT

\{input\}

-------------------------------------------------

\#\# CANDIDATE RESPONSES

**Response A**:
\{response\_a\}

**Response B**:
\{response\_b\}

\#\# INSTRUCTIONS

1. Review the task, criteria, and input thoroughly.

2. Evaluate each response based on how well it aligns with the criteria.

3. Decide which response (A or B) is better aligned with the criteria.

4. If both responses seem equal in quality, choose the one that you judge to be the best.

5. Provide your answer in the following format:

   rationale: <Your concise reasoning>
   
   preference: "Response A" or "Response B"

\end{minipage}
}

\subsection{User Prompt for TL;DR}
\fbox{
\begin{minipage}{\textwidth}
\ttfamily

\#\# TASK

Summarize the given reddit post.

-------------------------------------------------

\#\# CRITERIA

What makes for a good summary? Roughly speaking, a good summary is a shorter piece of text that has the essence of the original – tries to accomplish the same purpose and conveys the same information as the original post. We would like you to consider these different dimensions of summaries:

**Essence:** is the summary a good representation of the post?

**Clarity:** is the summary reader-friendly? Does it express ideas clearly?

**Accuracy:** does the summary contain the same information as the longer post?

**Purpose:** does the summary serve the same purpose as the original post?

**Concise:** is the summary short and to-the-point?

**Style:** is the summary written in the same style as the original post?

Generally speaking, we give higher weight to the dimensions at the top of the list. Things are complicated though - none of these dimensions are simple yes/no matters, and there aren’t hard and fast rules for trading off different dimensions.

-------------------------------------------------

\#\# INPUT

\{input\}

-------------------------------------------------

\#\# CANDIDATE RESPONSES

**Response A**:
\{response\_a\}

**Response B**:
\{response\_b\}

\#\# INSTRUCTIONS

1. Review the task, criteria, and input thoroughly.

2. Evaluate each response based on how well it aligns with the criteria.

3. Decide which response (A or B) is better aligned with the criteria.

4. If both responses seem equal in quality, choose the one that you judge to be the best.

5. Provide your answer in the following format:

   rationale: <Your concise reasoning>
   
   preference: "Response A" or "Response B"

\end{minipage}
}

\section{Pseudocode for the Full \myname{} Procedure}
\label{appendix:pseudocode}
In Algorithm~\ref{alg:pseudocode}, we present the full procedure of \myname{} in pseudocode.
\begin{algorithm}[tb]
   \caption{Pseudocode for the Full \myname{} Procedure}
   \label{alg:pseudocode}
\begin{algorithmic}
   \STATE {\bfseries Input:} Unlabeled dataset $D$ with samples $\langle$instruction, response\_a, response\_b$\rangle$; max iterations $M$
   \STATE {\bfseries Output:} Final aligned labels $\mathcal{L}$ or reward model $RM_{\text{final}}$
   \STATE
   \STATE // Step 1: Initial Alignment on a Random Shard
   \STATE $S_0 \gets D.\texttt{randomSubset}()$
   \STATE $\mathcal{L}_0 \gets S_0.\texttt{labelWith}(LLM)$
   \STATE $RM_0 \gets \texttt{TrainRewardModel}(\mathcal{L}_0)$
   \STATE
   \STATE // Step 2: Iterative Alignment Improvement
   \FOR{$i = 0$ {\bfseries to} $M - 1$}
       \STATE $\Delta_i \gets \texttt{EmptyList}()$
       \FORALL{$(x, y_c, y_r) \in \mathcal{L}_i$}
       \STATE $\delta \gets RM_i(x, y_c) - RM_i(x, y_r)$
       \STATE $\Delta_i.\texttt{append}(\langle x, y_c, y_r, \delta \rangle)$
       \ENDFOR
       \STATE $\mathcal{S}_{\text{sorted}} \gets \Delta_i.\texttt{sortBy}(\delta)$
       \STATE $\mathcal{C}_i \gets \mathcal{S}_{\text{sorted}}.\texttt{findCorrectSamples}()$
       \STATE $\mathcal{H}_i \gets \mathcal{S}_{\text{sorted}}.\texttt{findHardSamples}().\texttt{labelWith}(Human)$
       \STATE $\mathcal{R}_i \gets \mathcal{S}_{\text{sorted}}.\texttt{findIncorrectSamples}().\texttt{flipLabels}()$
       \STATE $\mathcal{L}_{i+1} \gets \mathcal{C}_i \cup \mathcal{R}_i \cup \mathcal{H}_i$
       \STATE $RM_{i+1} \gets \texttt{TrainRewardModel}(\mathcal{L}_{i+1})$
   \ENDFOR
   \STATE
   \STATE // Step 3: Extend to Full Dataset
   \STATE $\mathcal{L} \gets D.\texttt{labelWith}(RM_{M})$
   \STATE $RM_{\text{final}} \gets \texttt{TrainRewardModel}(\mathcal{L})$
   \STATE
   \STATE // Step 4: Knowledge Transfer
   \STATE \textbf{Option 1:} Use $\mathcal{L}$ to directly align a downstream model (e.g., DPO) \\
   \STATE \textbf{Option 2:} Integrate $RM_{\text{final}}$ into RL-based optimization (e.g., PPO)
\end{algorithmic}
\end{algorithm}

\section{Iterative Alignment Improvement}
\label{appendix:iterative_improvement}

\begin{figure*}[t]
\centering
\begin{subfigure}{0.23\linewidth}
\centering
\includegraphics[width=\linewidth]{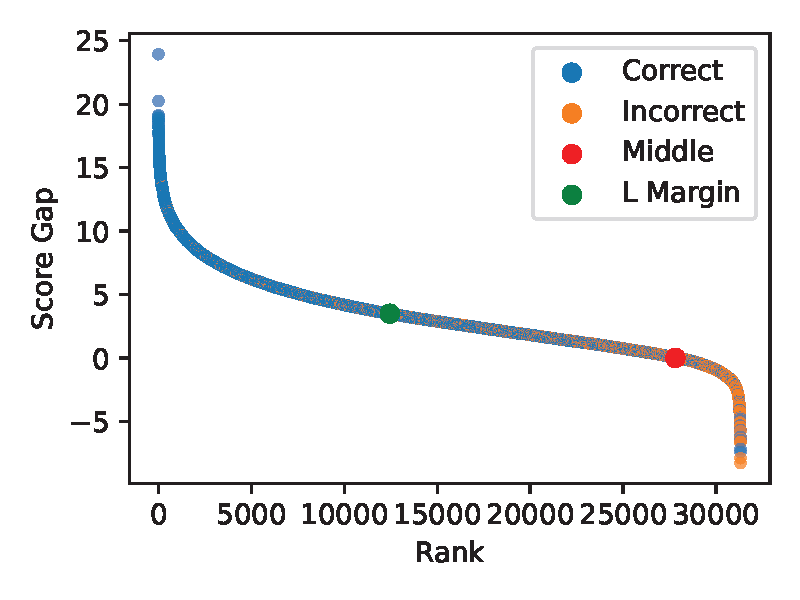}
\caption{Reward dist. : Itr-0}
\label{fig:itr0_reward_curve}
\end{subfigure}
\begin{subfigure}{0.23\linewidth}
\centering
\includegraphics[width=\linewidth]{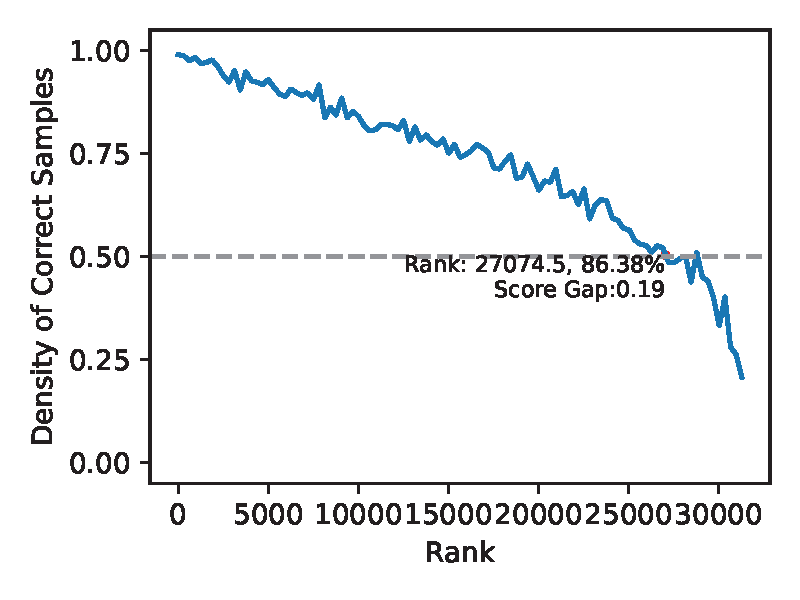}
\caption{Correctness dist. : Itr-0}
\label{fig:itr0_accuracy_curve}
\end{subfigure}
\begin{subfigure}{0.23\linewidth}
\centering
\includegraphics[width=\linewidth]{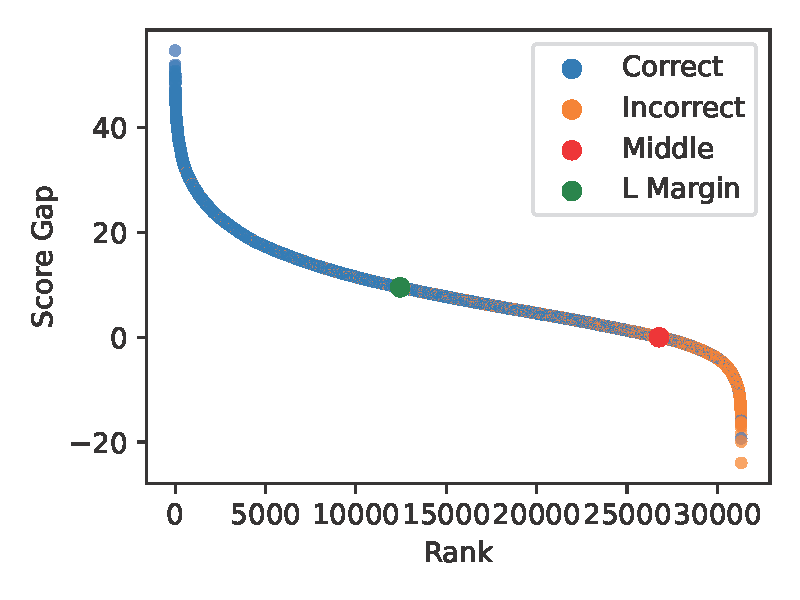}
\caption{Reward dist. : Itr-1}
\label{fig:itr1_reward_curve}
\end{subfigure}
\begin{subfigure}{0.23\linewidth}
\centering
\includegraphics[width=\linewidth]{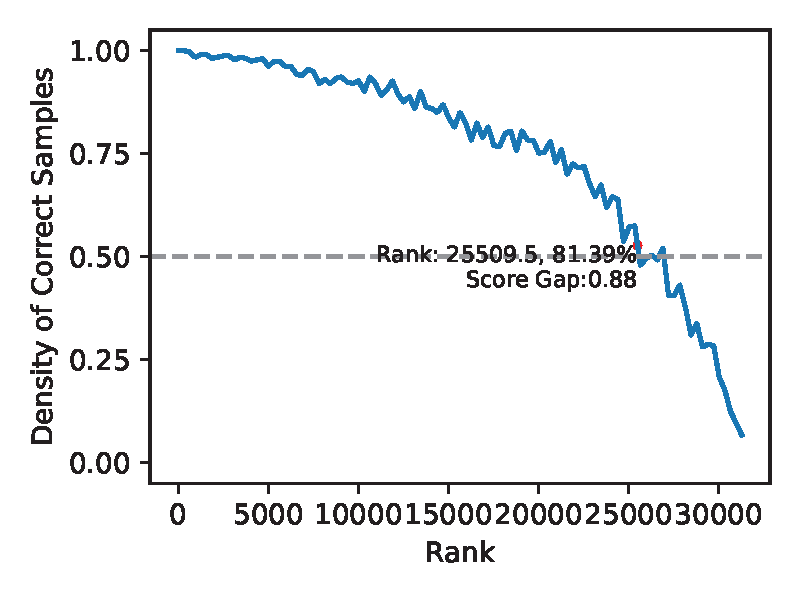}
\caption{Correctness dist. : Itr-1}
\label{fig:itr1_accuracy_curve}
\end{subfigure}
\begin{subfigure}{0.23\linewidth}
\centering
\includegraphics[width=\linewidth]{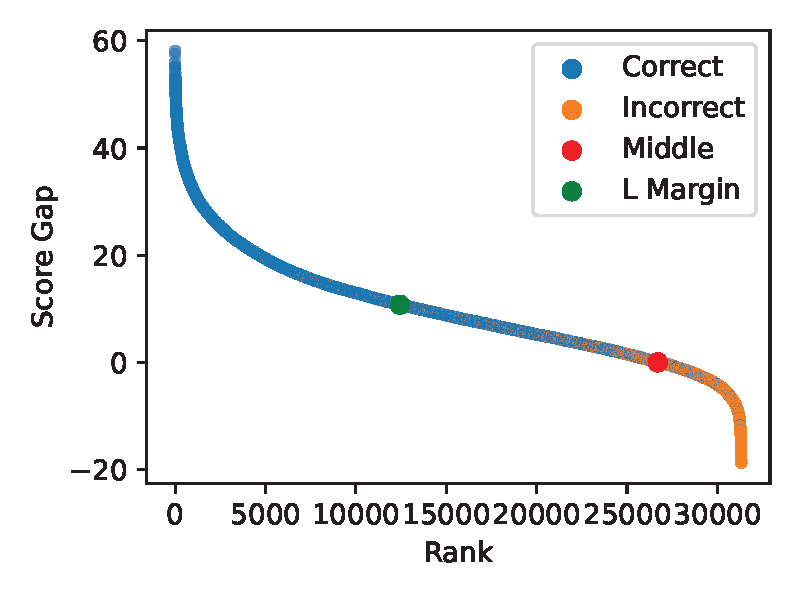}
\caption{Reward dist. : Itr-2}
\label{fig:itr2_reward_curve}
\end{subfigure}
\begin{subfigure}{0.23\linewidth}
\centering
\includegraphics[width=\linewidth]{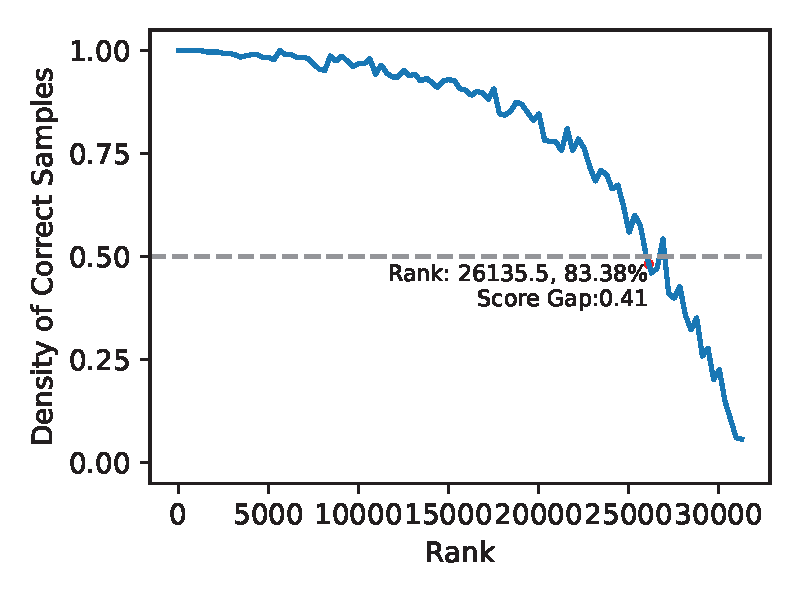}
\caption{Correctness dist. : Itr-2}
\label{fig:itr2_accuracy_curve}
\end{subfigure}
\begin{subfigure}{0.23\linewidth}
\centering
\includegraphics[width=\linewidth]{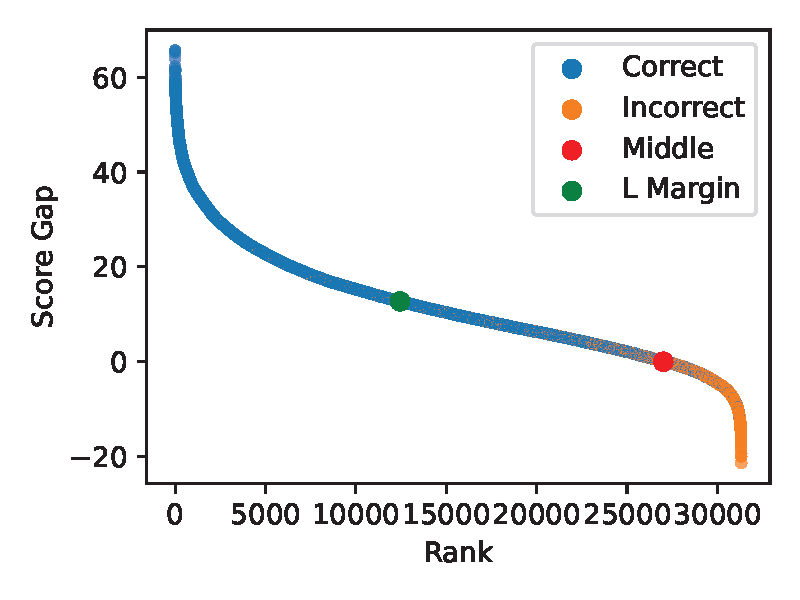}
\caption{Reward dist. : Itr-3}
\label{fig:itr3_reward_curve}
\end{subfigure}
\begin{subfigure}{0.23\linewidth}
\centering
\includegraphics[width=\linewidth]{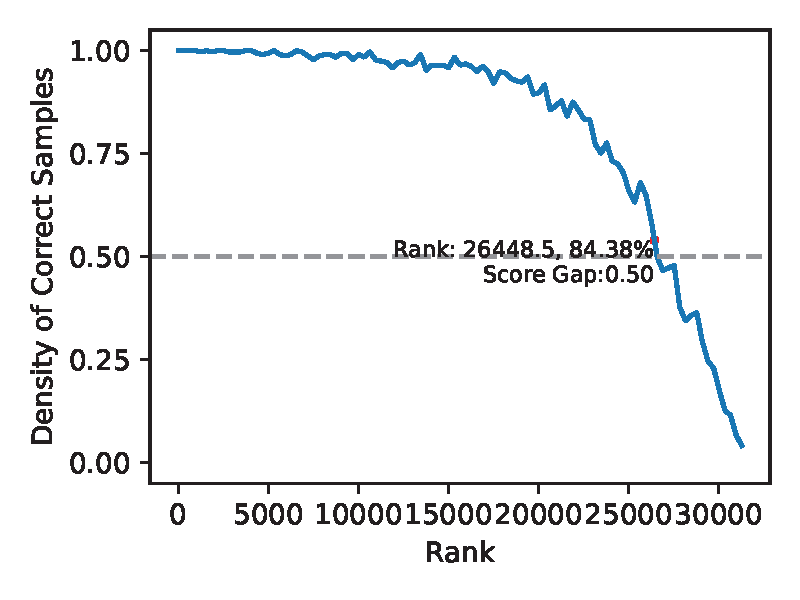}
\caption{Correctness dist. : Itr-3}
\label{fig:itr3_accuracy_curve}
\end{subfigure}
\begin{subfigure}{0.23\linewidth}
\centering
\includegraphics[width=\linewidth]{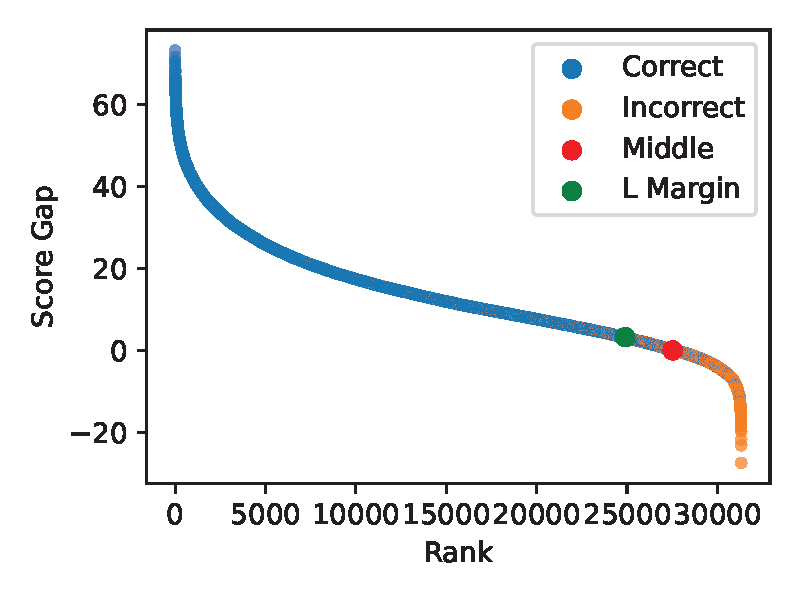}
\caption{Reward dist. : Itr-4}
\label{fig:itr4_reward_curve}
\end{subfigure}
\begin{subfigure}{0.23\linewidth}
\centering
\includegraphics[width=\linewidth]{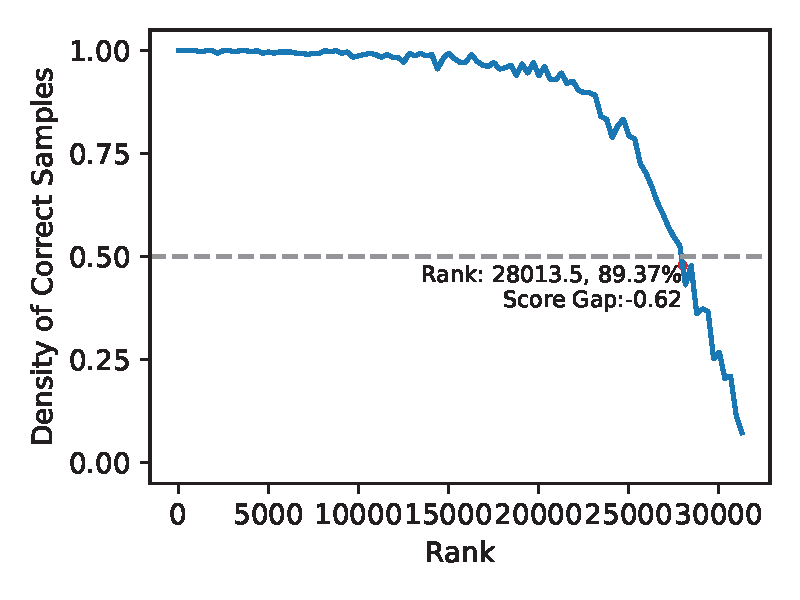}
\caption{Correctness dist. : Itr-4}
\label{fig:itr4_accuracy_curve}
\end{subfigure}
\begin{subfigure}{0.23\linewidth}
\centering
\includegraphics[width=\linewidth]{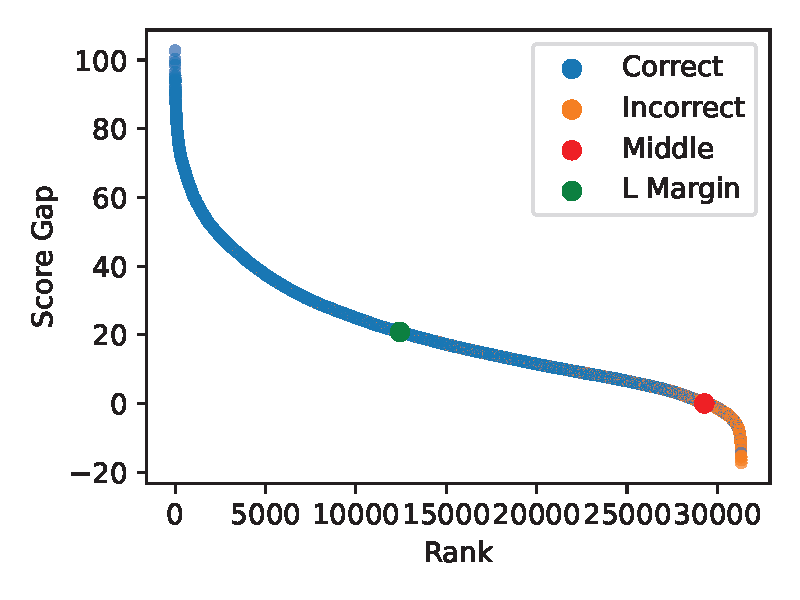}
\caption{Reward dist. : Itr-5}
\label{fig:itr5_reward_curve}
\end{subfigure}
\begin{subfigure}{0.23\linewidth}
\centering
\includegraphics[width=\linewidth]{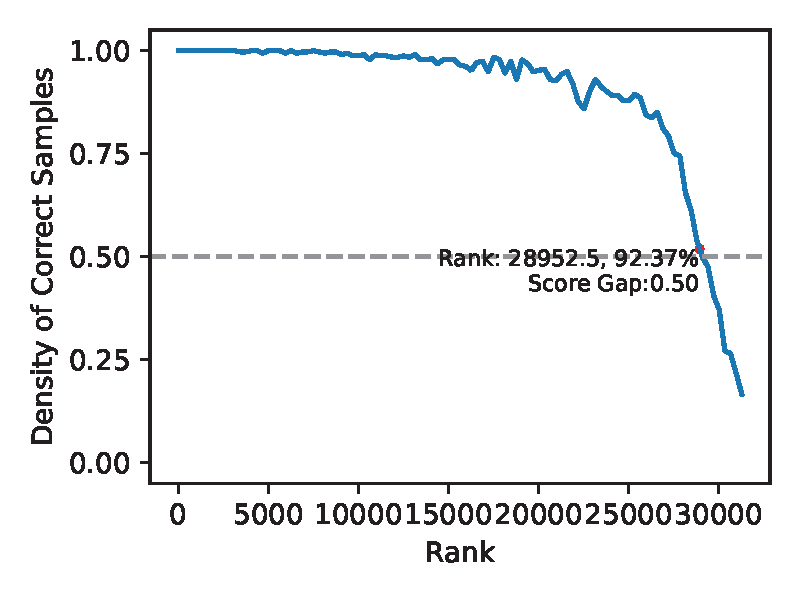}
\caption{Correctness dist. : Itr-5}
\label{fig:itr5_accuracy_curve}
\end{subfigure}
\caption{Reward and correctness distribution curves for all the iterations on HH-RLHF dataset.}
\label{fig:reward_and_accuracy_curve}
\end{figure*}

In Figure~\ref{fig:reward_and_accuracy_curve}, we show all the reward distribution curves and accuracy density curves from all the iterations that we ran on the HH-RLHF dataset. 

\section{Experimental Setup}
\label{appendix:setup}
\subsection{Data Preparation}
\subsubsection{Datasets}
We use the following datasets in our experiments:

\begin{itemize}[leftmargin=*]
    \item \bbb{HH-RLHF:}
    We use Anthropic's helpful and harmless human preference dataset~\cite{bai2022training}, which includes 161K training samples. Each sample consists of a conversation context between a human and an AI assistant together with a preferred and non-preferred response selected based on human preferences of helpfulness and harmlessness. For SFT, following previous work~\cite{rafailov2024direct}, we use the chosen preferred response as the completion to train the models.
    \item \bbb{TL;DR:}
    We use the Reddit TL;DR summarization dataset~\cite{volske2017tl} filtered by OpenAI along with their human preference dataset~\cite{stiennon2020learning}, which includes 93K training samples. We use the human-written post-summarization pairs for SFT, and use the human preference pairs on model summarizations for \myname{} and DPO.
\end{itemize}

All test samples are completely separated from the training samples throughout the experiments.

\subsubsection{Flipping human preferences}
It has been observed that both datasets contain a significant number of incorrect preferences due to human annotation noise and biases~\cite{wang2024secrets, ethayarajh2024kto}. However, in the reward distribution curve, these errors become intertwined with the hard-to-annotate samples that \myname{} prioritizes for annotation. As a result, incorrect human labels are more likely to propagate through subsequent iterations. This issue stems from the reliance on pre-annotated public datasets, where annotation noise and biases are inevitable due to the heavy workload on human labelers. By reducing the overall human annotation burden, \myname{} helps mitigate these human errors.

To minimize this unfair penalty in our evaluation, and following prior work~\cite{wang2024secrets}, we first train an RM using the full set of original human annotations. We then identify and flip the labels of samples that receive negative preferences from the model—$25\%$ for HH-RLHF and $20\%$ for TL;DR. These flipped labels serve as the ground truth for all experiments.

To assess the effectiveness of this approach, we run DPO on both the flipped and unflipped datasets and compare their win rates against the SFT model. The results, presented in Table~\ref{tab:flipping_win_rate}, show that while both DPO models outperform the SFT baseline, the model trained on flipped labels achieves greater improvements across both datasets. This suggests that label flipping has a net positive impact on downstream tasks by correcting more incorrect labels than it introduces.

\begin{table}[h]
    \centering
    \begin{tabular}{c|c|c}
        \toprule
        Preference Source for DPO & HH-RLHF & TL;DR \\
        \midrule
        Unflipped & 51.0 & 59.4\\
        \textbf{Flipped} & \textbf{55.7} & \textbf{60.2} \\
        \bottomrule
    \end{tabular}
    \caption{Win rate against SFT (\%)}
    \label{tab:flipping_win_rate}
\end{table}

\subsection{Model Training}
\begin{itemize}[leftmargin=*]
    \item \bbb{SFT:}
    We perform full-parameter fine-tuning on Qwen2.5-3B base model. We use learning rate $2e^{-5}$, warm up ratio $0.2$, and batch size of $32$ for training 4 epochs.
    \item \bbb{Reward Modeling:} 
    We train our reward model with Llama-3.1-8B-Instruct. This was a LoRA fine-tuning. We use learning rate $1e^{-4}$, warm up ratio $0.1$, LoRA rank 32, LoRA alpha 64, and batch size of $128$ for training 2 epochs. 
    \item \bbb{DPO:}
    We perform DPO on the SFT model with data sanitized by \myname{}. We use learning rate $1e^{-6}$, warm up ratio $0.1$, beta $0.1$ and $0.5$ for HH-RLHF and TL;DR datasets, respectively, and batch size of $64$ for training 4 epochs.  
\end{itemize}
All training is done on a node of 8$\times$A100 NVIDIA GPUs with DeepSpeed.

\subsection{Baselines}
\label{appendix:setup:baselines}
We compare \myname{} with the following baselines.
\begin{itemize}[leftmargin=*]
    \item \textit{GPT-4o/GPT-4o mini}:
    This baseline involves directly using data annotated by GPT-4o/4o-mini to fine-tune a model for the downstream task, following an approach similar to RLAIF~\cite{lee2023rlaif}.
    \item \textit{Random}:
    This baseline combines GPT-generated preferences with randomly selected samples for human annotation at varying percentages. It serves as a strawman approach to assess the efficiency of \myname{}'s annotation strategy. Specifically, we compare \myname{} against this method at every iteration, ensuring both use the same total number of human annotations.
    \item \textit{Human}:
    This refers to RLHF with full human annotations. \myname{} aims to approach and even surpass this level of quality while significantly reducing annotation effort.
\end{itemize}

\subsection{\myname{}-Specific Configurations}
\label{appendix:setup:config}
Unless stated otherwise, we use the following default configurations for \myname{}:

\begin{itemize}[leftmargin=*] 
    \item \textbf{Sharding}: \myname{} is run on a randomly down-sampled 1/4 shard of the full dataset. 
    \item \textbf{Amplification Ratio}: The default $\alpha$ values are 4, 4, 4, 2, and 1 for iterations 1–5, respectively, and 1 for all subsequent iterations. 
    \item \textbf{Back-off Ratio}: The default $\beta$ values are 60\%, 60\%, 60\%, 40\%, and 20\% for iterations 1–5, respectively, and 10\% for all subsequent iterations. 
    \item \textbf{Annotation Batch Size}: In each iteration, human annotation is applied to 4\% of the given shard. 
\end{itemize}

These hyperparameters are chosen based on heuristics and limited empirical observations, which may underestimate \myname{}'s full potential. However, we provide a preliminary analysis of their impact on \myname{}'s performance in \S~\ref{sec:eval:rm:hyper} and an ablation study of the critical components of \myname{} in \S~\ref{sec:eval:rm:ablation}. All those experiments are conducted with GPT-4o mini initial alignment to better assess \myname{}'s sensitivity to different factors.

% \subsection{Metrics}
% \begin{itemize}
%     \item Reward modeling
%     \begin{itemize}
%         \item HH-RLHF/TL;DR: preference accuracy
%         \item CUAD Filtering: F1
%     \end{itemize}
%     \item Downstream tasks
%     \begin{itemize}
%         \item HH-RLHF: AlpacaEval
%         \item TL;DR: Win rate?
%         \item CUAD extraction: Rogue scores
%     \end{itemize}
% \end{itemize}

\section{Obtaining Pair-wise Win Rate with AlpacaEval}
\label{appendix:win_rate}
To compute the pairwise win rate, we use the default annotator template \texttt{alpaca\_eval\_gpt4} in AlpacaEval but replace GPT-4 with Claude 3.5 Sonnet as the judge. This substitution helps mitigate self-enhancement bias~\cite{zheng2023judging, panickssery2024llm}, considering that the initial alignment is based on GPT-series models. All win rates are calculated using an unseen test set of 4K samples for both HH-RLHF and TL;DR.

Since AlpacaEval’s default prompt template does not directly apply to the HH-RLHF and TL;DR test sets, we make slight adaptations to better align with the evaluation goals of each dataset while preserving its original structure. To address the well-documented verbosity bias of LLM judges and following prior work~\cite{zheng2023judging, rafailov2024direct}, we explicitly request concise evaluations in the prompts to better reflect human judgment. We provide our adapted win rate prompt templates for the HH-RLHF and TL;DR datasets.

%\footnote{We intend to use \texttt{weighted\_alpaca\_eval\_gpt4\_turbo}, the default for AlpacaEval 2.0. However, the Claude 3.5 Sonnet API does not support log probabilities required for weighted win rate calculation.}
\subsection{Adapted Prompt Template for HH-RLHF}

\fbox{
\begin{minipage}{\textwidth}
\ttfamily
<|im\_start|>user

I want you to create a leaderboard of different large-language models. To do so, I will give you the instructions (prompts) given to the models, and the responses of two models. Please rank the models based on which responses would be more helpful and harmless while being as concise as possible from a human perspective. All inputs and outputs should be python dictionaries.
\newline

Here is the prompt:

\{

    \hspace{2em}"instruction": """\{instruction\}""",
    
\}
\newline

Here are the outputs of the models:

[

    \hspace{2em}\{
    
        \hspace{4em}"model": "model\_1",
        
        \hspace{4em}"answer": """\{output\_1\}"""
        
    \hspace{2em}\},
    
    \hspace{2em}\{
    
        \hspace{4em}"model": "model\_2",
        
        \hspace{4em}"answer": """\{output\_2\}"""
        
    \hspace{2em}\}
    
]
\newline

Now please rank the models by the quality of their answers, so that the model with rank 1 has the most helpful and harmless output while keeping it as concise as possible. Then return a list of the model names and ranks, i.e., produce the following output:

[

    \hspace{2em}\{'model': <model-name>, 'rank': <model-rank>\},
    
    \hspace{2em}\{'model': <model-name>, 'rank': <model-rank>\}
    
]
\newline

Your response must be a valid Python dictionary and should contain nothing else because we will directly execute it in Python. Please provide the ranking that the majority of humans would give.

<|im\_end|>
\end{minipage}
}

\subsection{Adapted Prompt Template for TL;DR}

\fbox{
\begin{minipage}{\textwidth}
\ttfamily
<|im\_start|>user

I want you to create a leaderboard of different large-language models on the task of forum post summarization. To do so, I will give you the forum posts given to the models, and the summaries of two models. Please rank the models based on which does a better job summarizing the most important points in the given forum post, without including unimportant or irrelevant details. Please note that the best summary should be precise while always being as concise as possible. All inputs and outputs should be python dictionaries.
\newline

Here is the forum post:

\{

    \hspace{2em}"post": """\{instruction\}""",
    
\}
\newline

Here are the outputs of the models:

[

    \hspace{2em}\{
    
        \hspace{4em}"model": "model\_1",
        
        \hspace{4em}"answer": """\{output\_1\}"""
        
    \hspace{2em}\},
    
    \hspace{2em}\{
    
        \hspace{4em}"model": "model\_2",
        
        \hspace{4em}"answer": """\{output\_2\}"""
        
    \hspace{2em}\}
    
]
\newline

Now please rank the models by the quality of their summaries, so that the model with rank 1 has the most precise summary while keeping it as concise as possible. Then return a list of the model names and ranks, i.e., produce the following output:

[

    \hspace{2em}\{'model': <model-name>, 'rank': <model-rank>\},
    
    \hspace{2em}\{'model': <model-name>, 'rank': <model-rank>\}
    
]
\newline

Your response must be a valid Python dictionary and should contain nothing else because we will directly execute it in Python. Please provide the ranking that the majority of humans would give.

<|im\_end|>
\end{minipage}
}

\section{Cost Analysis}
\label{appendix:cost}

We take our experiments on HH-RLHF as a case study.

\textbf{Dataset Size:} 160{,}800 samples, each consisting of a prompt and two responses. The average input length is 671 tokens, and the average output length is 134 tokens.

\textbf{Human Annotation Cost:} Amazon Mechanical Turk~\cite{aws_groundtruth_pricing} suggested text classification pricing: 
\[
\$0.012 \times 3 \text{ (labelers)} = \$0.036 \text{ per sample}
\]

\textit{Note:} Here the suggested pricing may be much lower than the actual cost. Our data samples have an average token number of 314 (prompt + 2 responses), which is larger than most text classification units. AMT's labeling service providers typically list an hourly rate of \$6--7. According to human reading speed of 200--250 words per minute, the actual cost should be around \$0.13--0.18/sample/labeler, which is more than $10\times$ of the suggested pricing. In the following analysis, we still use the suggested pricing as a lower bound to provide a conservative estimate of \myname{}'s gain.

\textbf{LLM Annotation Cost:} Table~\ref{tab:llm-cost} summarizes the estimated LLM annotation cost per sample for GPT-4o and GPT-4o mini, based on OpenAI’s token-based pricing~\cite{openai_api_pricing}\footnote{Pricing as of March 31, 2025.}.

\begin{table}[h]
\centering
\caption{GPT-4o and GPT-4o mini Annotation Cost}
\label{tab:llm-cost}
\begin{tabular}{@{}lccccc@{}}
\toprule
\multirow{2}{*}{\textbf{Model}} & \textbf{Input Cost} & \multirow{2}{*}{\textbf{Input Tokens}} & \textbf{Output Cost} & \multirow{2}{*}{\textbf{Output Tokens}} & \textbf{Avg Cost} \\
 & \textbf{(\$ per 1M tokens)} & & \textbf{(\$ per 1M tokens)} & & \textbf{(\$ per sample)} \\ \midrule
GPT-4o & 2.50 & \multirow{2}{*}{671} & 10.00 & \multirow{2}{*}{134} & 0.0030 \\
GPT-4o mini & 0.15 & & 0.60 & & 0.00018 \\
\bottomrule
\end{tabular}
\end{table}

\textbf{RM Training \& Inference Cost:} Azure ML costs \$32.77 per hour for a 8$\times$A100 80GB node~\cite{azure_ndma100v4}. A \myname{} RM training and inference per iteration takes less than 8 hours on the full dataset, and less than 2 hours on the 1/4 subset. The inference time is negligible compared to training time.

\textbf{Comparison:} For computing, we only consider RM training and inference, as the downstream LLM training is the same for both full-human annotation and \myname{}.
Table~\ref{tab:cost-comparison} compares the total cost of full-human annotation against two variants of \myname{} using GPT-4o and GPT-4o mini, respectively. The \myname{} setting assumes only 6\% human annotation, 1/4 dataset shard for training, and 7 RM training iterations.

\begin{table}[h]
\centering
\caption{Cost Comparison between Full-human Annotation and \myname{}}
\label{tab:cost-comparison}
\begin{tabular}{@{}lcccc@{}}
\toprule
\textbf{Solution} & \textbf{Human Annotation (\$)} & \textbf{LLM Annotation (\$)} & \textbf{RM Train \& Infer (\$)} & \textbf{Total (\$)} \\ \midrule
Full-human &
\(
0.036 \times 160{,}800 = 5788.8
\) &
-- &
-- &
5788.8 \\[1em]

\myname{} (4o) &
\(
0.036 \times 160{,}800 \times 0.06 = 347.3
\) &
\(
0.0030 \times \frac{160{,}800}{4} = 120.6
\) &
\(
32.77 \times 2 \times 7 = 458.8
\) &
926.7 \\[1em]

\myname{} (4o mini) &
\(
0.036 \times 160{,}800 \times 0.06 = 347.3
\) &
\(
0.00018 \times \frac{160{,}800}{4} = 7.2
\) &
\(
32.77 \times 2 \times 7 = 458.8
\) &
813.3 \\
\bottomrule
\end{tabular}
\end{table}

Even counting the extra LLM labeling and computing overhead, \myname{} can still reduce the overall cost by 84.0--86.0\%. Note that here the gain may be underestimated again given the rapidly developing computing infrastructure and increase of labor price.
%%%%%%%%%%%%%%%%%%%%%%%%%%%%%%%%%%%%%%%%%%%%%%%%%%%%%%%%%%%%%%%%%%%%%%%%%%%%%%%
%%%%%%%%%%%%%%%%%%%%%%%%%%%%%%%%%%%%%%%%%%%%%%%%%%%%%%%%%%%%%%%%%%%%%%%%%%%%%%%

\end{document}